\documentclass[letterpaper]{article}
\usepackage{aaai17}
\usepackage{times}
\usepackage{helvet}
\usepackage{courier}
\usepackage{booktabs} 
\usepackage{amsmath}
\usepackage{subcaption}
\usepackage{xcolor}
\usepackage{graphicx}
\usepackage{multirow}
\usepackage{comment}
\usepackage{url}

\newcommand{\octo}{\textsc{Octopus}}
\newcommand{\oct}{\textsc{Octopus}}
\newcommand{\controller}{\textsc{TaskSelector}}
\newcommand{\quality}{\textsc{QualityManager}}
\newcommand{\cost}{\textsc{CostSetter}}

\newcommand{\ff}{{\sc FrontierFinding}}

\newcommand{\vbar}{\bar{\nu}}
\newcommand{\bhat}{\theta}
\newcommand{\etal}{\mbox{\it et al.}}
\newcommand{\shorten}[1]{} 

\title{Octopus: A Framework for Cost-Quality-Time Optimization in Crowdsourcing}
\author{Karan Goel\thanks{Most work was carried out when the authors were students at the Indian Institute of Technology - Delhi.}\\
Carnegie Mellon University\\
kgoel93@gmail.com
\And
Shreya Rajpal$^*$\\
Univ. of Illinois at Urbana-Champaign\\
shreya.rajpal@gmail.com
\And
Mausam\\
Indian Institute of Technology - Delhi\\
mausam@cse.iitd.ac.in}

\begin{document}
\maketitle
\begin{abstract}
We present \octo, an AI agent to jointly balance three conflicting task objectives on a micro-crowdsourcing marketplace -- the quality of work, total cost incurred, and time to completion.
Previous control agents have mostly focused on cost-quality, or cost-time tradeoffs, but not on  directly controlling all three in concert. A naive formulation of three-objective optimization is intractable; \octo\ takes a hierarchical POMDP approach, with three different components responsible for setting the pay per task, selecting the next task, and controlling task-level quality.
We demonstrate that \octo\ significantly outperforms existing state-of-the-art approaches on real experiments.  We also deploy \octo\ on Amazon Mechanical Turk, showing its ability to manage tasks in a real-world, dynamic setting.
\end{abstract}

\section{Introduction}
Task control of workflows over micro-task crowdsourcing platforms, such as Amazon Mechanical Turk (AMT), has received significant attention in AI literature  \cite{weld2015handbook}. Typically, a requester needs to balance three competing objectives -- (1) total {\em cost}, owing to payments made to workers for their responses (or \emph{ballots}), (2) overall {\em quality}, usually evaluated as accuracy of the final output, and (3) the total {\em time} for completing the task. These criteria are inter-related: increasing the pay per task attracts more workers to the task, thereby reducing completion time. However, it  also exhausts the budget sooner, so requesters can afford fewer ballots per task, likely reducing the overall quality. 

Most prior work on crowd controllers has focused on the tradeoff between cost (or no. of ballots) and quality \cite{dai2013pomdp,Lin2012CrowdsourcingCM,bragg2013crowdsourcing,kamar2013lifelong,parameswaran2012crowdscreen}. A common approach is to define a Partially Observable Markov Decision Process (POMDP) {\em per task}, which decides on whether to get another ballot or submit the best answer for that task. However, this work is time-agnostic, and assumes that pay per ballot is given as input.

Recent work has also studied the tradeoff between cost and completion time for a {\em batch} of tasks \cite{gao2014finish}. They model the problem as a Markov Decision Process (MDP) that changes the pay per ballot, so that all ballots can be obtained by the given deadline in a cost-efficient manner. However, this work assumes that the number of ballots needed to complete the whole batch is a constant known to the requester in advance.

There is limited research on simultaneously addressing tradeoffs between cost, quality and latency. We know of only one work that studies this for the specific workflow for finding  {\em max} of a set of items \cite{venetis2012max}. {This work assumes that latency is pay-independent -- an assumption well-known to be incorrect \cite{faradani2011s,gao2014finish}. Under a fixed latency-per-response assumption, they speed up the task and save cost by taking fewer responses. }
Our three-way optimization is for the broader case of answering a batch of tasks, and uses variable pricing to alter the latency of task completion, in line with crowdsourced marketplace dynamics.


Building upon these strands of research, we present \octo, an AI agent that can balance all three objectives (cost, quality, time) in concert on real crowdsourced marketplaces, by optimizing a requester-specified, \emph{joint} utility function for a batch of tasks. It achieves this by controlling both the pay per ballot and the (predicted) accuracy of each individual task.


We could model the whole problem as a {\em single} POMDP, however, that is unlikely to scale. An alternative could be to use multi-objective MDPs, but they are also less tractable, because they produce a pareto-optimal set of solution policies 
\cite{chatterjee2006markov}. \octo\ uses a three-component architecture -- one to set the pay per ballot (\cost), another to choose the next available task (\controller) and a third to control each task's quality (\quality). A key technical novelty is in the careful modeling of the \cost's state space in order to circumvent intractability -- the state space contains aggregate statistics regarding completion levels of all tasks, so that it can decide the next best pay to set. 

We perform extensive experiments using both simulated and real data, as well as online experiments on AMT. Since no existing system performs direct 3-way optimization in crowdsourced marketplaces, our experiments compare against existing state-of-the art approaches that optimize 2 of the 3 objectives. We find that in most settings, \octo\ \emph{simultaneously} outperforms, or is at par with \emph{multiple} variants of these baselines. Our contributions are:
\begin{enumerate}
\item We describe \octo, a novel framework to address cost-quality-time optimization for a batch of tasks in a crowdsourced marketplace setting. It contains three components that set the pay per ballot, select the next task and control each task's quality. A key technical novelty is the use of aggregate statistics of all tasks in the state space design for the \cost, ensuring tractability for real-time deployment. 
\item \octo\ consistently performs at par with or better than state-of-the-art baselines, yielding up to 37\% reward improvements on real data.
\item We deploy \octo\ on AMT and demonstrate that it is able to optimize utility effectively in a live, online experiment.\footnote{Code can be found at \url{https://github.com/krandiash/octopus}.}
\end{enumerate}

\section{Related Work}

\noindent 
{\bf Cost-Quality Optimization.} There is significant work on getting more quality out of a fixed budget. One branch of this research focuses on {\em collective classification}, which develops aggregation mechanisms to infer the best output per task, given a static set of ballots  \cite{whitehill2009whose,welinder2010multidimensional,oleson2011programmatic,welinder2010online}.
The other branch studies {\em intelligent control}, which dynamically decides whether to ask for a new ballot on a task, or stop and submit the answer. 
These include control of binary or multiple choice tasks \cite{dai2013pomdp,parameswaran2012crowdscreen,kamar2013lifelong}, multi-label tasks \cite{bragg2013crowdsourcing}, and tasks beyond multiple choice answers \cite{Lin2012CrowdsourcingCM,Dai2011ArtificialIF}. All these works design agents to control a {\em single} task and assume a constant pay per ballot.
Our work closely follows the POMDP formulation laid down in Dai \etal\ \shortcite{dai2013pomdp} for binary tasks. 
\nocite{kamar2013lifelong}

\noindent {\bf Cost-Time Optimization.} Increasing pay per ballot can reduce completion times. Faradani \etal\ \shortcite{faradani2011s} develop models to find upfront, the static price per ballot so that a desired deadline can be met. Gao \& Parameswaran \shortcite{gao2014finish} extend this by varying pay at discrete time-steps using an MDP. Both approaches assume a fixed number of ballots known a-priori, without dynamic quality control of tasks. There is also some work on price-independent latency reduction \cite{haas2015clamshell}.

\noindent {\bf Cost-Quality-Time Optimization.} There is limited work in this area. The only paper we are aware of is Venetis \etal\ \shortcite{venetis2012max}, which addresses cost-quality-time optimization but in a restrictive setting with important distinctions from our work: (i) they look at max-finding for a set of items, while our task-type is classification; (ii) they consider latency to be pay-independent and fixed per task, while we study the more realistic setting in which changing pay directly impacts workers' desire to work on our tasks; (iii) unlike us, they \emph{don't} change pay per task directly, and instead, change the number of responses sought per task to control both cost and latency. Qualitatively, our work thus also highlights how workers perceive pay changes in a crowdsourcing marketplace and its overall effect on task completion.

\noindent
{\bf Worker Retention.} Previous work deals with incentivizing workers to perform more tasks via bonuses or diversification  \cite{rzeszotarski2013inserting,ipeirotis2014quizz,difallah2014scaling,dai2015and}. Recently, Kobren \etal\ \shortcite{kobren2015getting} model the process of worker retention. We present empirical results that suggest worker retention plays a dominant role in determining task completion rates.

\noindent 
{\bf Task Routing.} Prior work deals with two issues; deciding which task from the batch to solve next, or which worker to route a task to.  Ambati \etal\ \shortcite{ambati2011towards} rank tasks based on user preferences using a max-entropy classifier. Other work uses low-rank matrix approximations \cite{karger2014budget} for equal difficulty tasks. 
Rajpal \etal\ \shortcite{rajpalpomdp} decide which worker pool to route a task to. Other papers study task routing on volunteer platforms \cite{bragg2014parallel,shahaf2010generalized}. 

Like AMT, we assume no control on which worker picks a ballot job, but we select the best next task to assign to an incoming worker.
Following \cite{mason2010financial,gao2014finish}, we assume that worker quality is independent of the pay per ballot. We re-verify this for our data in our experiments.

\noindent
{\bf Decentralized Approaches.} There is related work in decentralized Wald stopping problems \cite{teneketzis1987decentralized} which considers how to optimize a common utility function given a set of agents who each make independent observations. However, these approaches do not scale well with the number of agents (tasks in our setting), which can be quite large. There is also work in decentralized metareasoning \cite{hansen2001monitoring} to decide when to stop optimizing a utility function. Metareasoning approaches typically assume that utility is monotonically increasing over time, which is not true in our setting since for instance, conflicting ballots on a single task would decrease utility.

\section{Problem Definition}
A requester provides a batch of $n$ binary tasks $q\in 1\dots n$, each having a $0/1$ response. They also provide a utility function ${\cal U}$, which describes how to tradeoff cost, time and quality. The agent can dynamically change pay per ballot $c$, and choose a variable number of ballots per task to optimize the final objective. We study the setting where ${\cal U}$ is expressed as a sum of task-level utilities ($U$) minus cost, \emph{i.e.} ${\cal U} = \sum_{q=1}^n (U_q-C_q)$. Here, $C_q$ is total money spent on $q$.  We assume that answers to all tasks are to be returned to the requester as {\em one} single batch.

\noindent {\bf MDP/POMDP background.} An MDP models the long-term reward optimization problem under full observability and is defined by a five tuple $\langle S, A, T, R, \gamma \rangle$. Here, $S$ is a set of states, $A$ a set of actions, $T(s'|s,a)$ denotes the probability of transitioning to state $s'$ after taking action $a$ in state $s$, and $R(s,a)$ maps a state-action pair to a real-valued reward. $\gamma$ is the discount factor for making infinite-horizon MDPs well formed. A POMDP extends an MDP into a partially-observable setting, where the state is not fully observable and only a belief (probability distribution) over possible states can be maintained using observations from the model. A POMDP is represented as $\langle S, A, T, R, O, \gamma \rangle$ tuple, where a new function $O(o|s',a)$ denotes the distribution over observations on taking an action $a$ and arriving in a new state $s'$. Lack of space precludes a long discussion of the subject -- there are existing solvers for solving MDPs and POMDPs of reasonable sizes, e.g. \cite{smith2012point}, which we use in our work.

To formulate the problem optimally, we would need to define a single, centralized POMDP over a state containing answers and difficulty estimates of all individual tasks as well as the current pay per ballot and the current time. The actions will include requesting a ballot on a task $q$, changing the pay, and a terminal submit action. Solving this POMDP would yield the optimal policy, which would decide which task to get ballots on next, when to change pay and when to submit. Since the number of tasks in a batch can be huge, this POMDP is unlikely to scale due to a large state and action space. Naive extensions to Dai \etal 's or Gao \etal 's state-of-the-art models for cost-quality and cost-time optimization respectively are not possible either -- Dai \etal 's model is solved per task, whereas pay must be set based on progress of the whole batch of tasks; Gao \etal 's model assumes a fixed number of ballots per task, and has no natural way to optimize quality by taking a variable number of ballots based on each task's difficulty.

\begin{figure}[h!]
\centering
\includegraphics[width=1\linewidth]{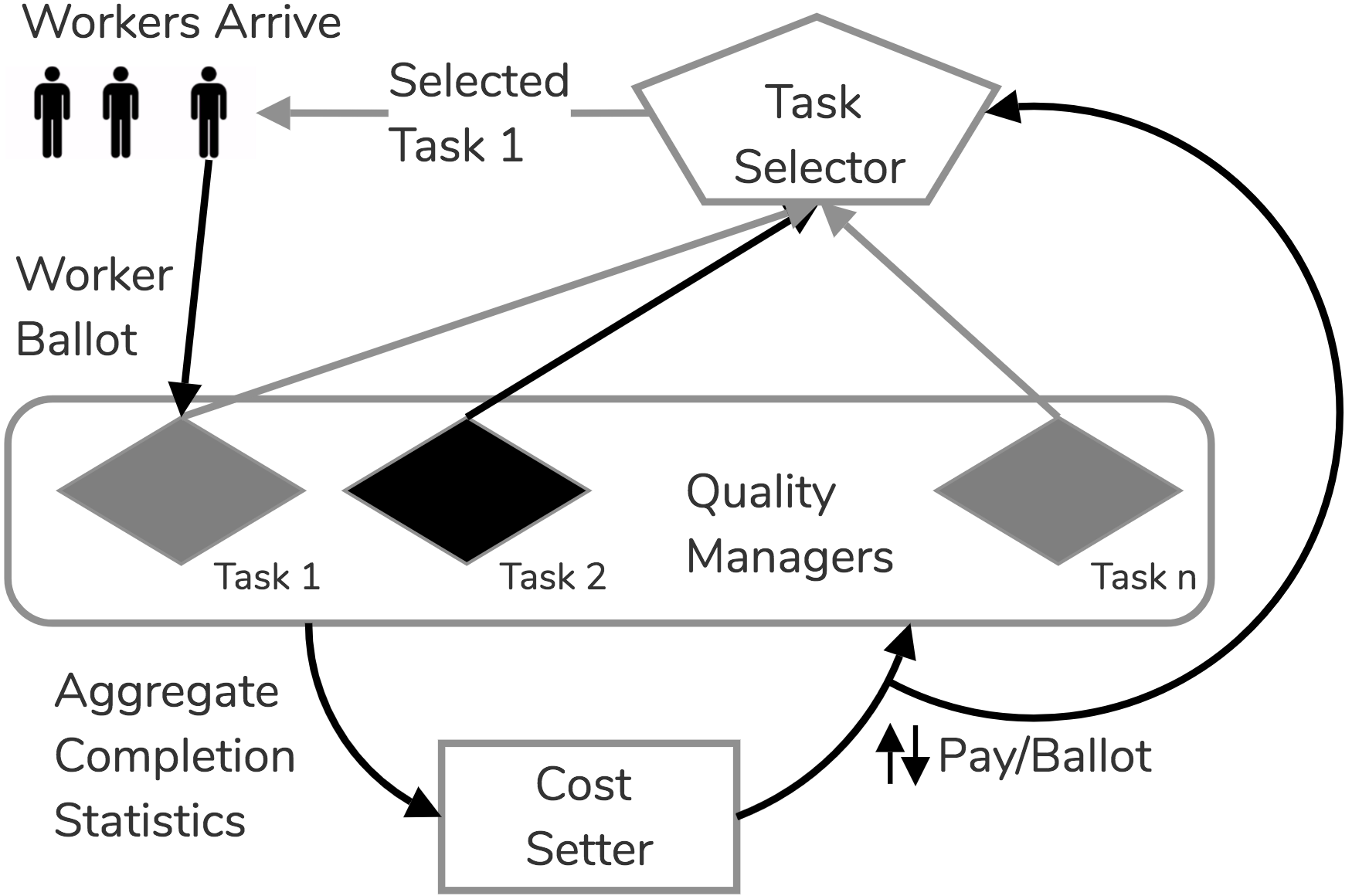}
\caption{
\octo\ architecture.}
\label{oct_block}
\end{figure}

\section{\octo\ for Three-Way Optimization}
We propose a three-component architecture (see Figure \ref{oct_block}). In \octo\, each task has its own \quality\ that decides, based on the current pay, whether it is worth taking another ballot for this task (light edge) or not (dark edge). This information is conveyed to the \controller, which selects an available (light) task to route to an incoming worker. Based on the current progress of the whole batch, the \cost\ decides what pay per ballot to set; this action is taken at periodic intervals. We design \octo\ such that it can allocate tasks as and when workers arrive and works instantaneously in practice, therefore utilizing crowdsourcing marketplaces with full parallelism. 

\subsection{QualityManager}

\noindent{\bf Background.}
\quality s follow the worker response model and POMDP formulation of Dai \etal\ \shortcite{dai2013pomdp}. Each worker is assumed to have an error parameter $\gamma\in (0,\infty)$ ($\gamma = 0$ is error-free). An average worker has error parameter $\bar{\gamma}$. Each task $q$ has an unknown true Boolean answer $t_q$, and an associated difficulty $d_q\in [0,1]$ ($d_q = 0$ is easy) -- these are estimated using an Expectation-Maximization algorithm \cite{whitehill2009whose} as data is received.
Each task has a prior difficulty distribution $p(d_q)$. 
A worker's ballot for $q$ depends on their $\gamma$, $t_q$ and $d_q$.

In Dai \etal\ \shortcite{dai2013pomdp} the POMDP for $q$ maintains a belief state $\mathbf{b}_q$ over $(d_q,t_q)$ state tuples.  
For instance $\mathbf{b}_q(0.5, 1) = 0.4$ indicates a 40\% belief that $d_q = 0.5$ and the answer to $q$ is $1$. 
The POMDP has two actions; (1) request another ballot, or (2) mark $q$ as completed. The POMDP policy $\pi_q$  maps every belief state to an action. Each time the POMDP receives another ballot, a Bayesian update is performed to re-estimate a new belief. For \octo, the POMDP optimizes $U_q-C_q$.
We note that $\pi_q$ depends on the pay per ballot $c$: if $c$ is smaller, the POMDP can afford more ballots. Dai \etal 's original model keeps $c$ constant, but in our case the \cost\ can change it, triggering a change in $\pi_q$.

We define $v_q$ as the confidence in the most probable answer for task $q$; $v_q = \mathrm{max}(v_q^0,v_q^1) \in[0.5,1]$, where $v_q^0,v_q^1$ are the current probabilities that $q$'s answer is 0 or 1 respectively. $v_q$ can be computed from $\mathbf{b}_q$ by summing out $d_q$. 

\noindent{\bf Computation of Aggregate Statistics.} We now define two batch-level statistics, which will be a key part of the state space representation for the \cost. 

We first define a normalized estimate of {\em task quality}, $\nu_q = 2v_q - 1$; $\nu_q$ normalizes $v_q$ so that it lies in $[0,1]$. A high value of $\nu_q$ (near 1) indicates the POMDP's high confidence in its estimated answer for $q$, and vice versa for a low (near 0) value. We also define a related notion of \emph{batch quality}, $\bar{\nu} = \frac{1}{n} \sum_{q=1}^{n} \nu_q$, which is an aggregate statistic estimating the current quality for the entire batch of $n$ tasks.
Finally, we construct the \emph{batch quality histogram} -- a histogram built by binning tasks into equally sized bins based on their $\nu_q$ values. The bin width is denoted as $\Delta_\nu$.

The \cost\ also needs an estimate of  the number of ballots remaining. We define $\theta_q(\pi_q)$ as an estimate of the expected number of ballots that will be needed (starting from the current time) until task $q$ will be marked completed. For notational ease we write $\theta_q$ to denote $\theta_q(\pi_q)$. Recall that $\pi_q$ can change with a change in $c$. Hence $\theta_q$ also depends on $c$.


How can we compute $\theta_q$? We use a trajectory-tree approach (similar to \cite{kearns1999approximate}) called \ff. We construct a binary tree of future observations, rooted at the current time step, where each edge corresponds to an observation (a worker response of $0/1$). The node below any edge contains the belief state generated by updating the POMDP, using the observation associated with that edge. Each trajectory is a path from the root to a leaf, and is generated with an associated path probability (using Dai \etal 's generative model assuming an average worker). A leaf is created whenever the policy takes the `mark as completed' action, or when the path probability drops below a threshold. $\theta_q$ is simply the expected length of a trajectory in this tree. 

We also estimate the expected ballots to completion for the batch: $\theta = \sum_{q=1}^n \theta_q$. $\theta$'s role is similar to that of $\bar{\nu}$ -- it is an aggregate statistic that describes how far the batch is from completion. It also helps in quantifying the expected cost of completion: if $\theta = 1000$ and $c=3$, we would expect to spend  $c\cdot\theta=3000$ units of money to complete the batch. 

In summary, we described the design of a per-task \quality. Collectively, $n$ of these help us in estimating two aggregated quantities, $\bar{\nu}$ and $\theta$, which measure the overall quality, and degree of completion of the batch, respectively. All notation for this and future sections is summarized in Table~\ref{notation_table}.



\subsection{TaskSelector}

The \controller\ decides which incomplete task to assign to the next incoming worker. 
It must have an `anytime' behavior, \emph{i.e.} it must increase utility ${\cal U}$ quickly. This is because 
the time of final submission is not in its control, and the batch might be submitted at any time by the \cost.



To be prepared for any contingency, the 
\controller\ uses a 1-step greedy policy over expected utility gain. We define each task's priority ($\phi_q$) as the difference between the current utility ($U_q$) of $q$ and the expected utility after receiving 1 ballot ($U'_q$) from an average worker (error rate $\bar{\gamma}$), given the current belief state $\mathbf{b}_q$ of $q$'s \quality. Thus, $\phi_q = \mathbf{E}[U_{q}'|\bar{\gamma},\mathbf{b}_q] - U_{q}$. 
\controller\ assigns the task with the maximum $\phi_q$ value to the next available worker.

Unfortunately, $U_q$ (and therefore ${\cal U}$) is {neither} monotonic (conflicting ballots decrease utility) nor submodular (a skilled worker could arrive after an error-prone one), so we cannot utilize prior work on adaptive submodularity \cite{golovin2011adaptive,bragg2014parallel} to guarantee solution quality. Providing quality bounds is left for future work. 

Lastly, note that the task allocation process is instantaneous, as well as completely parallelized, since we don't wait for a task to be returned before allocating another task. Given enough workers, we could get ballots on every single task in parallel. This is important, since it allows us to take full advantage of micro-task marketplaces.


\begin{table}[h!]
\centering
\begin{tabular}{|c|l|}
\toprule
\textbf{Symbol}   & \textbf{Meaning}                                                              
                                                    \\ \toprule
$\mathcal{U}$               & Overall utility (requester defined function)                \\ 
$U_q$               & Per-task utility                                                                                                   \\ 
$c$               & Pay per ballot  \\ 
$C$               & Total cost incurred for all tasks                                                                                                  \\ 
$C_q$               & Total cost incurred for task $q$                                                                                                  \\ 

$\pi_q$           & Policy for task $q$'s \quality\                                                                                                        \\ 
$\gamma$, $\bar{\gamma}$           & Worker error parameter, average error                                                                                                       \\ 
$d_q$             & Difficulty of task $q$                                                                                          \\ 
$t_q$             & True binary answer for task $q$                                                                                          \\ 
$\mathbf{b_q}$    & Belief state for task $q$                                                                                 \\ 
$v_q$             & \begin{tabular}[l]{@{}l@{}}Probability of POMDP's most likely true\\ \hspace{3mm}answer for task $q$\end{tabular}     \\ 
$\nu_q$           & \begin{tabular}[c]{@{}c@{}}Task quality ($\nu_q = 2v_q - 1$)\end{tabular}                   \\
$\bar{\nu}$       & \begin{tabular}[c]{@{}c@{}}Batch quality ($\bar{\nu} = \frac{1}{n}\sum_{q} \nu_q$)\end{tabular}               \\ 
$\theta_q$        & \begin{tabular}[l]{@{}l@{}}Estimated number of ballots required to\\ \hspace{3mm}complete task $q$\end{tabular} \\ 
$\theta$        & \begin{tabular}[l]{@{}l@{}}Estimated number of ballots required for\\ \hspace{3mm}completing the batch of tasks ($\theta = \sum_q \theta_q$)\end{tabular} \\ 
$\tilde{\theta}(\nu_q, c)$ & Mapping from $(\nu_q, c)$ to $\theta_q$ for task $q$ under $\pi$                                                                                      \\ 
$\phi_q$          & \begin{tabular}[c]{@{}c@{}}Priority of task $q$ \end{tabular}                                                                                                          \\
$\tau_{max}$      & Deadline                                                                                                      \\ 
$\tau$            & Current Time                                                                                                    \\
$\Delta$   & Granularity (\emph{e.g.} $\Delta_\tau$)                                                                                                                   \\ 
\hline
\end{tabular}
\caption{Notation used in the paper.}
\label{notation_table}
\end{table}

\subsection{CostSetter}
The \cost\ is an MDP that changes $c$ (pay per ballot) in order to maximize ${\cal U}$. It uses information about the completion level of each task to assess whether the batch of tasks is completing on schedule or needs to be sped up or slowed down. To influence the rate of completion of the batch, it sets $c$ at discrete time steps $\tau \in \{0,\Delta_\tau,2\Delta_\tau,\dots\}$.

The key challenge for the \cost\ is in defining the state space. Ideally, as stated earlier, each task's belief $\mathbf{b}_q$ should be part of the state, but that would make computations intractable. Instead, we approximate by using aggregate statistics over the whole batch of tasks. We describe the state space, actions, transition functions, and rewards of this MDP below.

\noindent {\bf State Space.} The choice of the best pay per ballot $c$ depends on its current value, the current time, and the aggregate degree of completion of the batch. We choose the state to be a 4-tuple $(\vbar,\bhat,\tau,c)$. {Both $\vbar$ and $\bhat$ are important for this decision; $\vbar$ estimates the expected accuracy on the batch, while $\theta$ gives us an idea of how much more improvement in $\vbar$ is possible (at the current $c$). For a fixed $\vbar$ and $c$, a high $\theta$ would indicate the presence of several unsolved tasks and the possibility of improving $
\vbar$. On the other hand, a low $\theta$ would indicate that most tasks are solved and there is little improvement possible. $\theta$ therefore captures the \emph{spread} of the distribution of task qualities $\nu_q$, while $\vbar$ is the mean of this distribution. We use this intuition later to construct the transition function for the \cost.}

{As another example consider a case where both $\theta$ and $\vbar$ are high. This indicates that the \quality s consider there be to scope for quality improvement despite the batch quality being high already, possibly due to very low $c$. If we did not have $\theta$ in the state, we would instead base our decision on the high $\vbar$ value, and believe that further improvement in utility was not possible.}

All state variables are continuous, and for tractability we discretize them. $\bhat$ is discretized with a granularity $\Delta_{\bhat}$, $\vbar$ with a granularity $\Delta_{\nu}$, and $\tau$ with a granularity $\Delta_\tau$. We assume that $c$ can take values $\{c_1,c_2\dots,c_k\}$. These values are defined by the requester, and in practice would respect marketplace constraints, such as minimum wages. Interestingly, the model is robust in that if a requester provides a very poor starting wage, workers will likely not pick up the task, and the model will subsequently respond by increasing the wage.



\noindent 
{\bf Actions and Rewards.} Every state has access to two pay-change actions: $\uparrow$ and $\downarrow$. $\uparrow$ increases $c_i$ to $c_{i+1}$ while $\downarrow$ does the opposite. The $\uparrow$ and $\downarrow$ actions incur no cost to the system. However, in practice we assign a small cost to these actions to prevent frequent cyclical pay-changes in a policy. Changing $c$ has no impact on $\tau$ and $\vbar$, but it does change $\bhat$, since the number of ballots remaining per task depends on the current pay per ballot (via the policy $\pi_q$). We discuss how to compute this when defining the transition function. 

We also have a no-change action, which increments $\tau$ by $\Delta_\tau$, along with asking workers for more ballots at $c$, which remains unchanged (for the next $\Delta_\tau$ duration). This is essentially a marketplace action, where we post tasks to the marketplace  with a pay per ballot equaling $c$. At the end of $\Delta_\tau$ minutes, we would then arrive in a new state. The cost of this transition (to the nearest $\Delta_{\theta}$ value) is just the amount paid to workers during this duration on the marketplace, equaling the number of ballots received during this time, multiplied by $c$.

The final action is a `terminate' action that submits all answers to the requester. Its reward should be ${\cal U}$ based on batch quality and current time (cost is not needed, since that was already accounted in the no-change action). Unfortunately, the MDP has access to only the aggregate statistics, and not the full batch quality histogram, which is needed for computing ${\cal U}$. We now describe a novel $\beta$-reconstruction procedure that allows us to extrapolate the full histogram from aggregate statistics, useful for computing this reward as well as the transition function.




\noindent{\bf $\mathbf{\beta}$-Reconstruction.} The goal is to reconstruct an approximate batch quality histogram given the aggregate statistics, $\vbar$ and $\bhat$, 
and the current $c$. The procedure assumes that the histogram can be approximated with a two parameter Beta distribution, $\beta_{\lambda_1, \lambda_2}$. Also assume that we are provided a function $\tilde{\theta}$ that maps a $(\nu_q, c)$ pair to $\theta_q$; it returns the expected number of ballots needed for a task $q$ given its current quality and pay per ballot. Note also that $\tilde{\theta}$ will be a non-increasing function of $\nu$. 
We now show that we can find suitable $\lambda_1$ and $\lambda_2$ given $\vbar$, $\bhat$ and $\tilde{\theta}$. 



To compute the best fit $\lambda_1,\lambda_2$ values, we solve two equations. The $1^{\mathrm{st}}$ equation enforces that the mean of the reconstructed distribution is $\vbar$: $\frac{\lambda_1}{\lambda_1 + \lambda_2} = \vbar$.
\shorten{
\begin{equation*}
\frac{\theta_1}{\theta_1 + \theta_2} = \vbar
\end{equation*}
}
We can reparameterize $\beta_{\lambda_1, \lambda_2}$ using $\lambda_1 = \lambda \vbar$; $\lambda_2 = \lambda (1 - \vbar)$ and write it as $\beta_\lambda(\vbar)$. Our task now reduces to finding the best fit $\lambda$. A $2^{\mathrm{nd}}$ equation imposes $\bhat$ as an expectation over the batch quality distribution: $\hat{\theta}(\lambda) = n \int_0^{1} \tilde{\theta}(\nu,c) \beta_\lambda(\nu) d\nu \approx \bhat$.

\shorten{
\begin{equation}
\label{eqn:ballots}
\bhat = n \int_0^{1} h_{\pi}(v',c) \beta_\lambda(v') dv'
\end{equation}
}
Since $\tilde{\theta}$ is computed using a POMDP policy, it will rarely be available in closed form. The integral above is approximated using a numerical algorithm. The best $\lambda$ is found via $\mathrm{argmin}_{\lambda} |\hat{\theta}(\lambda) - \bhat|$ using a linear search, and works instantaneously in practice. Having found a suitable $\lambda$ value, we now bin all $n$ tasks into the $\beta$ distribution to recover the task quality histogram, as desired.

Finally, we describe the procedure for estimation of $\tilde{\theta}(\nu_q,c)$. First we calculate the corresponding $v_q=0.5(\nu_q + 1)$. Intuitively, $\tilde{\theta}(\nu_q,c)$ assesses the number of ballots taken by the \quality\ when its belief $\mathbf{b}_q$ in the current answer is $v_q$ and it has difficulty $d_q$. However, we haven't reconstructed $d_q$ -- we use the prior distribution $p(d)$ as its belief on $d_q$. To compute $\tilde{\theta}(\nu_q,c)$ we initiate the \quality's POMDP from such a belief state, and compute $\bhat_q$ under policy $\pi_q$ using \ff.

\noindent{\bf $\uparrow/\downarrow$ Transitions.}
We now describe the transitions for $\uparrow$ action in a given state $(\vbar,\bhat,\tau,c_i)$ to reach $(\vbar,\bhat',\tau,c_{i+1})$. As we change $c$, the main change is in $\bhat$. We quantify this change by a simple observation: regardless of pay, the number of ballots taken until this point is fixed. Suppose that at $\tau=0$ we compute the estimated number of ballots for the batch as $\bhat_{0}(c_i)$. If we have taken $x$ ballots till now, our current estimate of $\bhat$ will be simply $\bhat_{0}(c_i) - x$. At pay $c_{i+1}$, our next state's estimate should be $\bhat_{0}(c_{i+1}) - x$. Thus, when changing pay from $c_i$ to $c_{i+1}$  we can simply add $\bhat_{0}(c_{i+1}) - \bhat_{0}(c_i)$ to compute $\theta'$. A similar analysis works for the $\downarrow$ action.

\noindent 
{\bf No-Change Transitions.} The no-change action emulates the setting that the tasks are posted on the platform for $\Delta_\tau$ time at pay $c$. Let the next state be  $(\vbar',\bhat',\tau+\Delta_\tau,c)$. Estimation of $\vbar'$ and $\bhat'$ requires a model of task completion. Similar to Gao \& Parameswaran \shortcite{gao2014finish}, we maintain a pay-dependent ballot completion model, $\mathrm{Pr}(n_b|\Delta_\tau,c)$, as the probability that \octo\ will receive $n_b$ ballots in duration $\Delta_\tau$ at pay $c$. However, different from their work, the probability model combines the effects of worker arrival, retention and time taken per task (and not just arrival). Thus, $\bhat$ will reduce by $n_b$ with probability $\mathrm{Pr}(n_b|\Delta_\tau,c)$. The cost of the transition will be $-c\cdot n_b$. Since $n_b$ is discretized upto $\Delta_{\bhat}$ granularity, the cost will be rounded off to the nearest bucket.

For updating $\vbar$, we $\beta$-reconstruct the batch quality histogram using the current state. We then bin $n$ tasks into this histogram, and simulate the \controller\ on this reconstructed batch. To do this, we first create a POMDP belief state for each reconstructed task by choosing $\nu$ from the histogram to recover $v$, and using the prior difficulty distribution. We then select a task, simulate a ballot using an average worker ($\bar{\gamma}$) and compute the posterior belief. We continue until all $n_b$ ballots are used up. At the end we compute the $\vbar'$ based on the updated state of the batch. For robustness, we repeat this entire procedure multiple times and average the $\vbar$ values from different runs.




\noindent 
{\bf Implementation Details.}
We construct the whole MDP with all transitions and rewards using simulations and $\beta$-reconstructions as described above. Since time can be unbounded, we keep a max time $\tau_{\max}$ when defining the total state space. 
We also recognize that the transition from $\vbar$ to $\vbar'$ depends on $n_b$ but not on any other part of the state. By caching a table of $(\vbar, n_b, \vbar')$ values once, we can save on a lot of simulations when computing the transition function of the no-change action in various states. 

We use Value Iteration to learn the policy with $(\vbar = 0, \bhat = n\cdot \max \tilde{\theta}(\nu = 0,c = c_1), \tau = 0, c = c_1)$ as our start state. Intuitively, we are starting at 0 quality, the maximum possible number of ballots to completion, and the lowest price.

In a real execution environment, it is possible that over time, the \cost's aggregates  diverge from the set of \quality's beliefs in the batch. To improve performance, we synchronize the \cost 's $\vbar$ and $\bhat$ to those of the real batch after every time interval. Experimentally \octo\ performs well even without synchronization, indicating that the learned transitions are effective.

In summary, we described the \cost, an MDP that keeps track of the aggregate statistics $\vbar$ and $\bhat$ for a batch of $n$ tasks, and changes pay to optimize utility ${\cal U}$. A key contribution is a novel $\beta$-reconstruction procedure that approximates the batch quality histogram for a state.

\begin{figure}[t!]
    \centering
    \begin{subfigure}{0.24\textwidth}
    \captionsetup{width=0.99\linewidth}
        \centering
        \includegraphics[width=\linewidth]{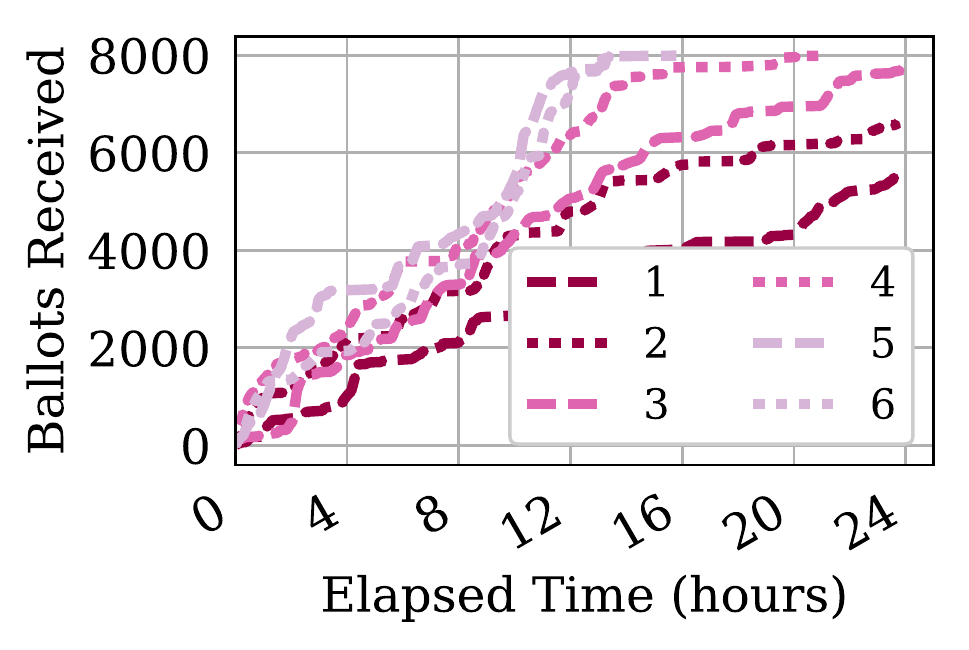}
        \label{fig:analysis_1}
    \end{subfigure}%
    \begin{subfigure}{0.24\textwidth}
	    \captionsetup{width=0.94\linewidth}
        \centering
        \includegraphics[width=\linewidth]{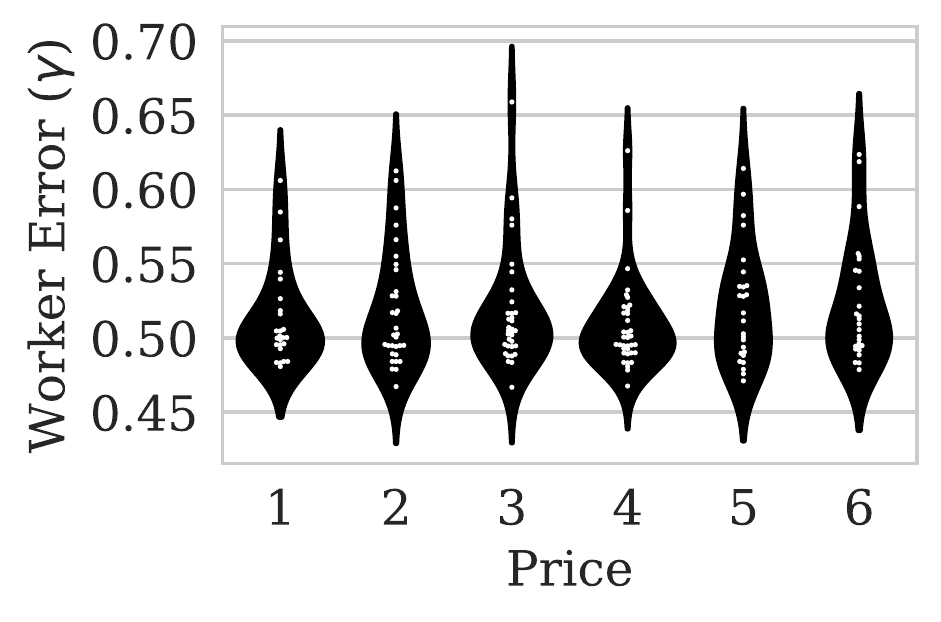}
                \label{fig:analysis_2}
    \end{subfigure}
    \caption{(a) Ballots received vs. elapsed time for different costs (in tenths of a cent); (b) Worker error distribution for different costs are similar.}
    \label{fig:leftover}
\end{figure}

\begin{figure}[b!]
    \centering
    \begin{subfigure}{0.24\textwidth}
    \captionsetup{width=0.99\linewidth}
        \centering
        \includegraphics[width=\linewidth]{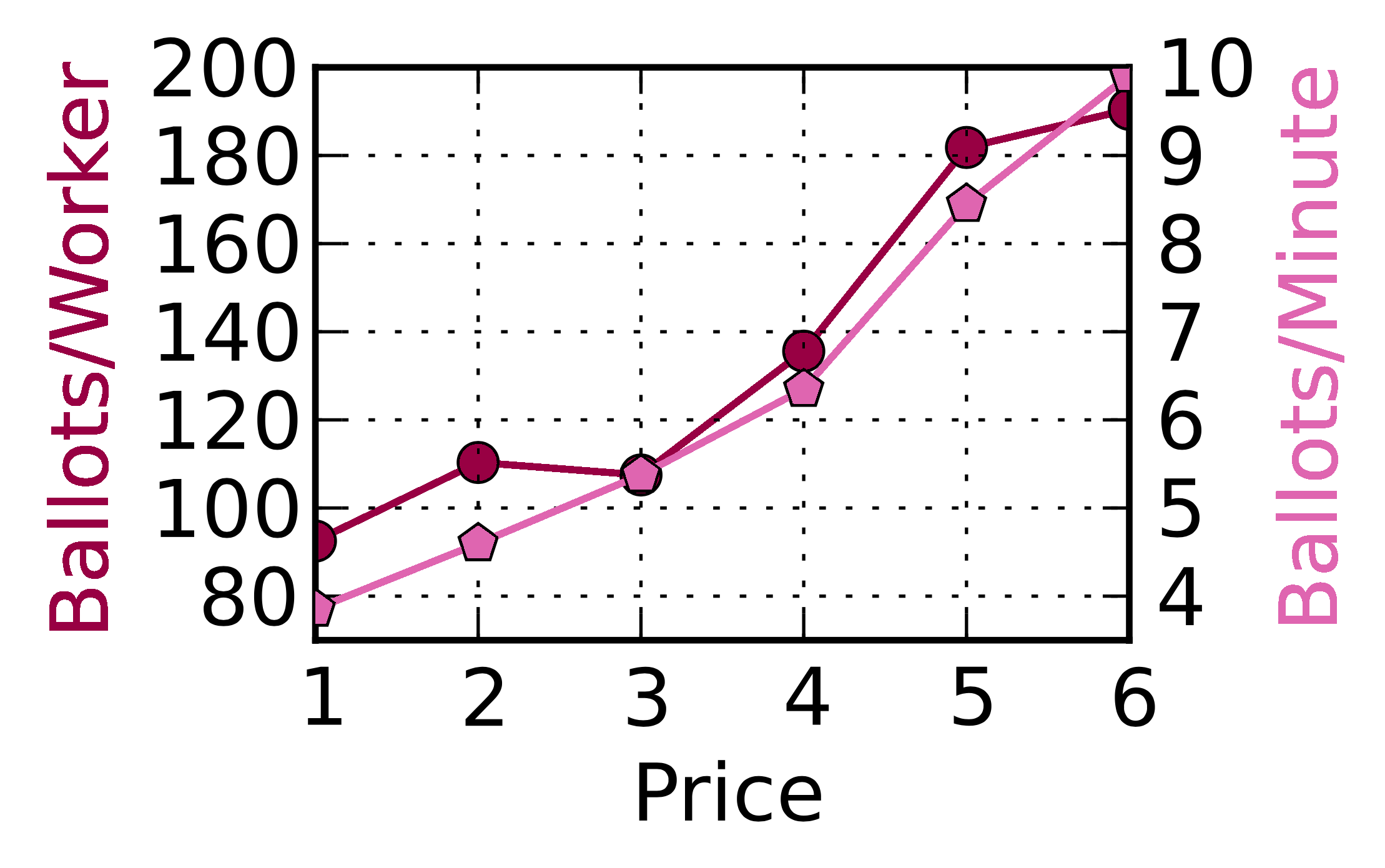}
        \label{fig:analysis_1}
    \end{subfigure}%
    \begin{subfigure}{0.24\textwidth}
	    \captionsetup{width=0.94\linewidth}
        \centering
        \includegraphics[width=\linewidth]{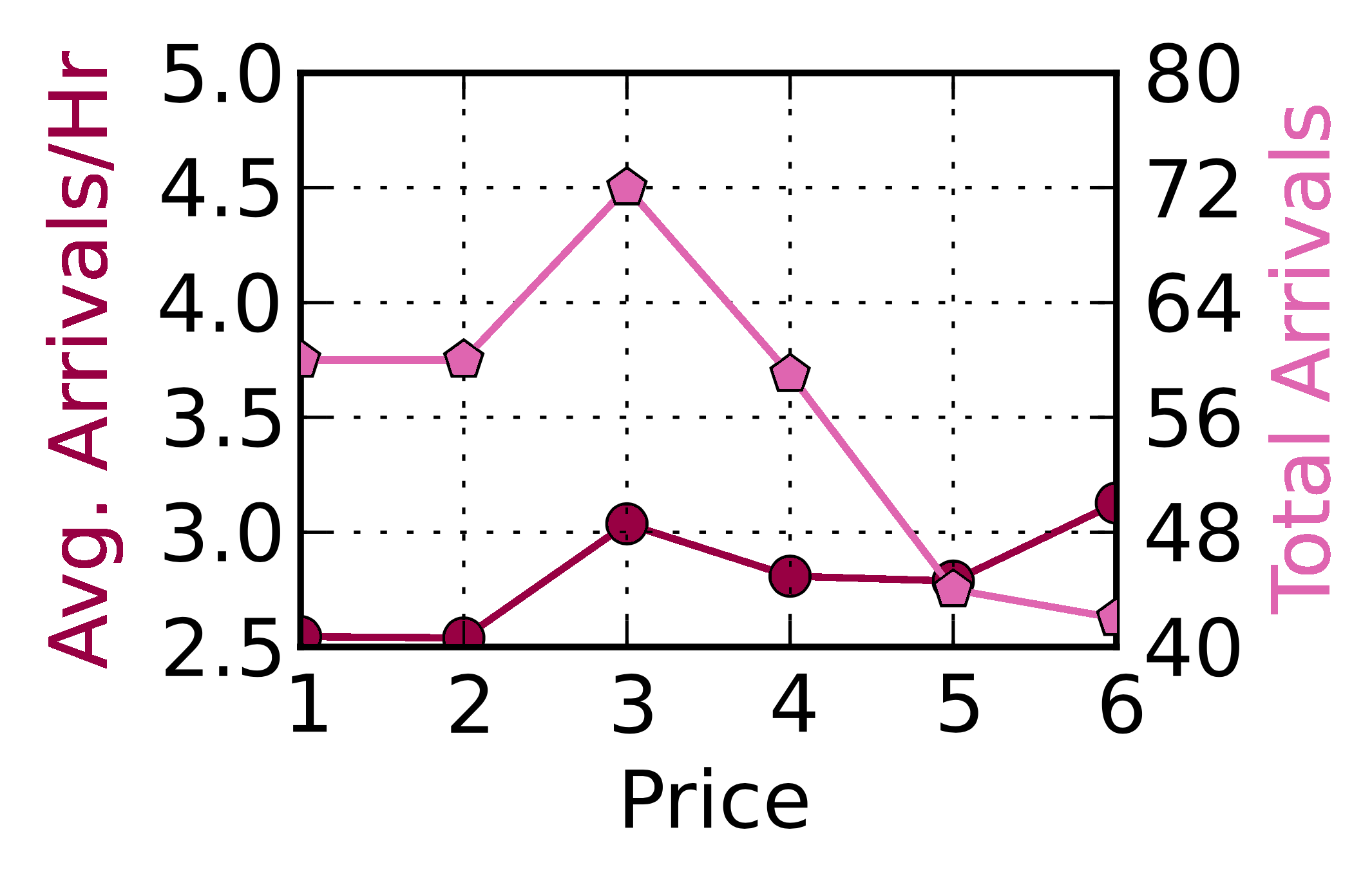}
                \label{fig:analysis_2}
    \end{subfigure}
    \caption{Task completion statistics for our data: (a) Retention \& Completion Rate; (b) Arrivals (Rate \& Total)}
    \label{fig:analysis}
\end{figure}

\section{Experiments}
We conduct three sets of experiments: (i) experiments on simulated data; (ii) offline evaluation on real data; (iii) live experiments on Amazon MTurk (AMT).

\noindent{\bf Model Parameters.} For experiments, we initialize \octo\ with a hard deadline utility function: $U_q = -\infty$ if $\tau < \tau_{\max}$; otherwise, $U_q=-{\cal P}$ for every incorrect answer and zero for a correct answer. This joint utility combines the utilities from Dai \etal 's and Gao \etal 's models. Here, penalty ${\cal P}$ represents how important quality is to the requester. Notice that since a POMDP doesn't know whether it is submitting the correct answer, it cannot compute the utility exactly. It uses its belief to estimate {\em expected} utility as $-{\cal P}(1-v_q)$. This linear utility makes the reward computations for \cost's `terminate' action simple. Given the $\vbar$ of the state, the reward is calculated as $-0.5{\cal P}(1-\vbar)$ on termination.

Having a hard deadline makes the \cost\ state space finite, since we only consider states with $\tau \le \tau_{\max}$. We use a time-independent Poisson process for the ballot completion model $\Pr(n_b|\Delta_\tau, c)$. These parameters are illustrative -- \octo\ can accommodate other utilities/distributions.

\noindent {\bf Baselines.}
Our main goal is to compare \octo's performance with state-of-the-art methods. However, 
we know of no algorithms that performs direct three-way optimization in a crowdsourced marketplace setting.\footnote{We don't compare to Venetis \etal\ \shortcite{venetis2012max} since they run a tournament for max-finding, different from our task type. They also don't change pay directly or model latency as done by us.} So
we compare against state-of-the-art methods that optimize 2 of the 3 objectives, while giving them the added benefit of our \controller. We compare \octo\ to Dai \etal 's and Gao \etal 's models, which are related to our work. Code was provided by the authors.

All comparisons are on the requester's utility function, ${\cal U}$ computed against gold labels with ${\cal P}=200$. We normalize ${\cal U}$ so that \octo\ always has 1.0 utility, \emph{i.e.} the performance of baselines is represented as a proportion of \octo.



\noindent {\bf Data collection.} We collect data on AMT at 6 pay points using a Twitter Sentiment dataset \cite{Sheshadri2013SQUAREAB}. Workers are asked to classify the sentiments of tweets into {either} positive or negative {sentiment}. At each price point $(\$0.001, \$0.002,\dots, \$0.006$ per ballot), we post \shorten{a batch of }400 tweets, and seek 20 ballots/tweet. A single \shorten{Human Information Task (}HIT contains 10 tweets for a worker to solve. 
40 tweets are common to all prices for a total of 2200 tweets. Each price is posted on a different weekday at the same time, and remains active for 24 hours. This ensures consistency in data collection across prices and minimizes interaction between different prices.

Figure~\ref{fig:leftover}a (task completion rates vs. pay) verifies that higher pricing results in faster \shorten{rate of}task completion: at 0.1 cent, only 5500 ballots are received \shorten{(out of 8000)}even after 24 hours, whereas at 0.6 cents, all 8000 ballots are received within 12 hours. \shorten{This verifies the basic premise of our framework, that higher pay leads to a faster task completion.  We were surprised to find a slightly  slower completion in 0.6 cent tasks compared to 0.5 cent ones, which we attribute to chance. \bug}We estimate the Poisson parameters using this data.

\noindent{\bf Worker Retention vs. Arrival.}
Contrary to prior work \cite{faradani2011s,gao2014finish}, we observe that the increase in task completion rate with pay is \emph{predominantly} due to higher worker \emph{retention}, rather than a higher rate or number of worker \emph{arrivals} {(possibly due to the large number of HITs that we sought)}. Figure~\ref{fig:analysis}a shows that both retention and task completion rates are highly correlated, doubling as price goes from 0.1 to 0.6 cents. \shorten{It is also clear from the figure that both quantities are highly correlated -- both roughly double as price goes from 1 to 6.}However, Figure~\ref{fig:analysis}b shows that the worker arrival rate does not rise much\shorten{with price}. 
To the best of our knowledge, there is no prior marketplace model that handles both retention and arrivals.

We also verify that worker quality is independent of pay. We run a K-S test for every pair of costs. We could not reject the null hypothesis (error rates drawn from cost-independent distributions) at $p<0.05$ (Figure \ref{fig:leftover}b).

\begin{figure*}[t!!]
	\centering
    \begin{subfigure}[t]{0.3\textwidth}
        \centering              
        \captionsetup{width=0.93\linewidth}
        \includegraphics[width=\linewidth, height=.7\linewidth]{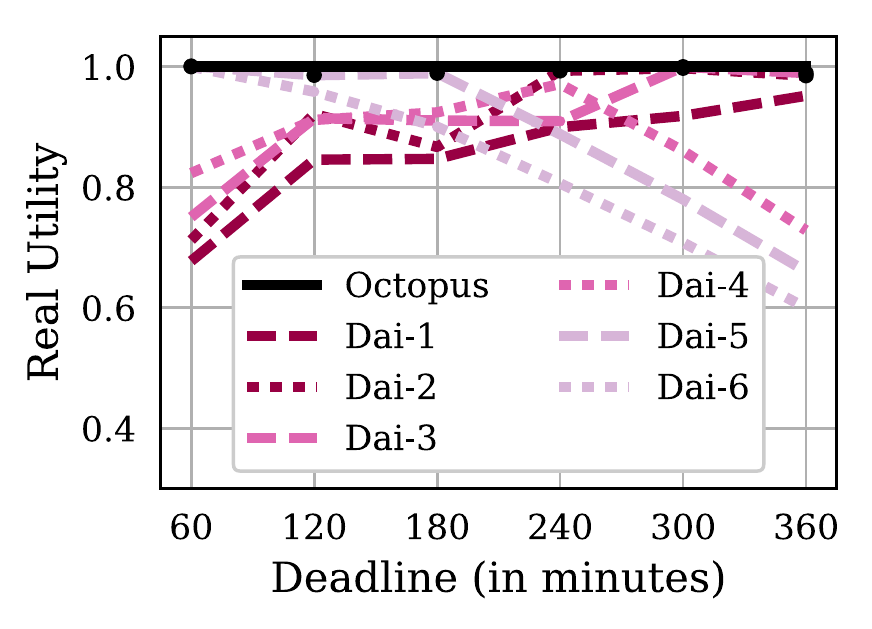}
        \caption{Comparison to \textsc{Dai}.}
        \label{fig:static_live}
    \end{subfigure}
    \begin{subfigure}[t]{0.3\textwidth}
        \centering
        \includegraphics[width=\linewidth, height=.7\linewidth]{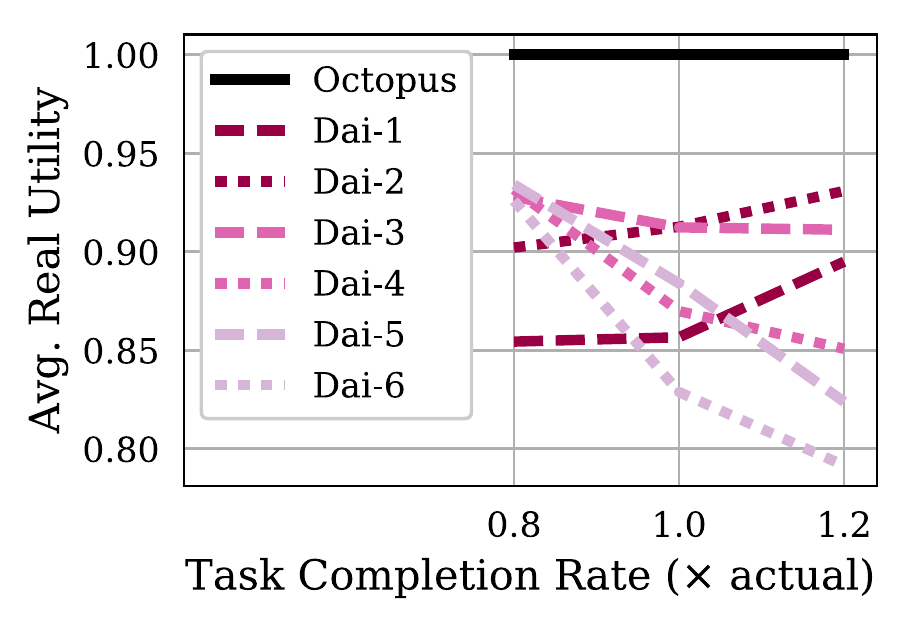}
        \captionsetup{width=0.93\linewidth}
        \caption{Comparison to \textsc{Dai}.}
        \label{fig:gao_live}
    \end{subfigure}%
        \begin{subfigure}[t]{0.3\textwidth}
        \centering
        \includegraphics[width=\linewidth, height=.7\linewidth]{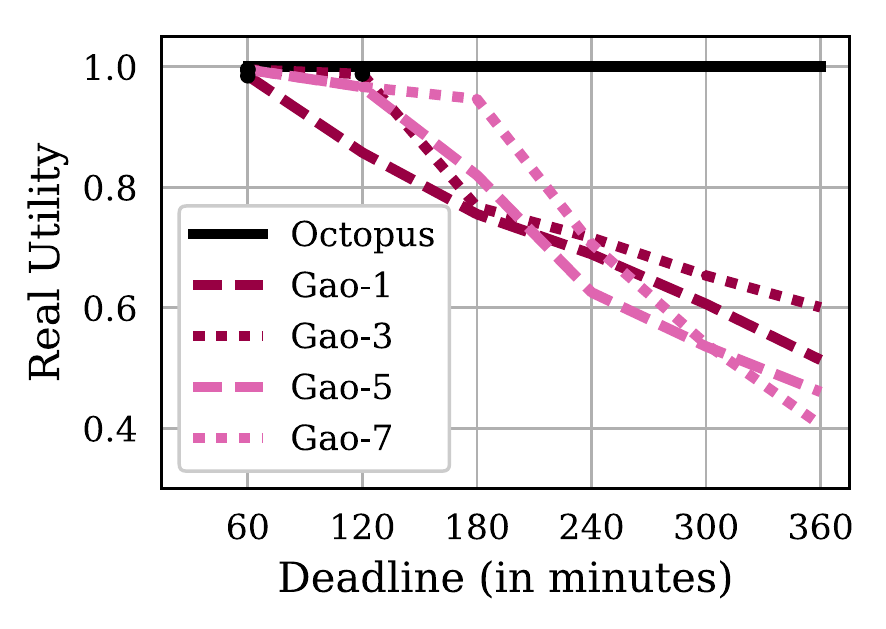}
        \captionsetup{width=0.93\linewidth}
        \caption{Comparison to {\sc Gao}.}
        \label{fig:controller_live}
    \end{subfigure}
    \begin{subfigure}[t]{0.25\textwidth}
        \centering              
        \captionsetup{width=0.93\linewidth}
        \includegraphics[width=\linewidth, height=.7\linewidth]{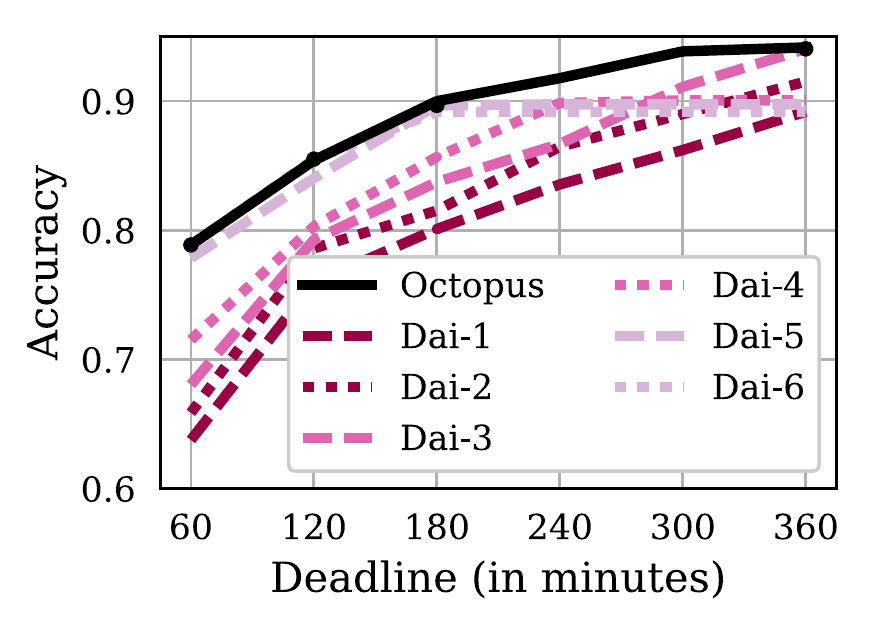}
        \caption{Comparison to \textsc{Dai}.}
        \label{fig:static_live}
    \end{subfigure}%
    \begin{subfigure}[t]{0.25\textwidth}
        \centering
        \includegraphics[width=\linewidth, height=.7\linewidth]{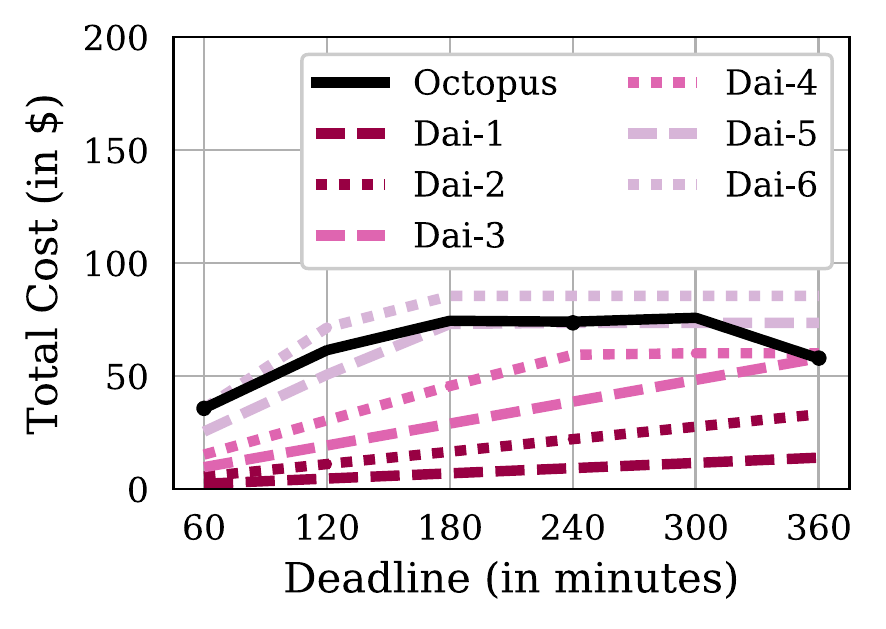}
        \captionsetup{width=0.93\linewidth}
        \caption{Comparison to \textsc{Dai}.}
        \label{fig:gao_live}
    \end{subfigure}%
        \begin{subfigure}[t]{0.25\textwidth}
        \centering
        \includegraphics[width=\linewidth, height=.7\linewidth]{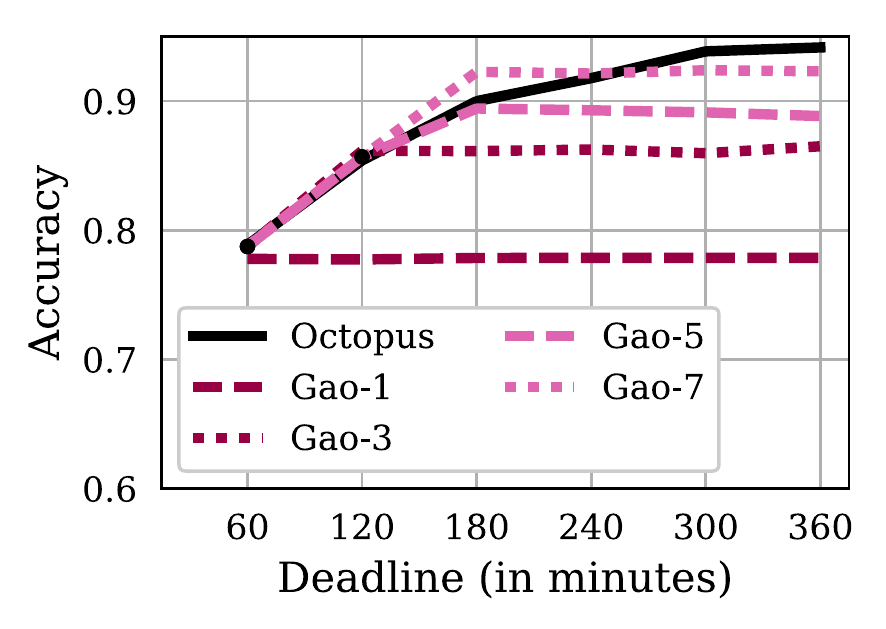}
        \captionsetup{width=0.93\linewidth}
        \caption{Comparison to {\sc Gao}.}
        \label{fig:controller_live}
    \end{subfigure}%
    \begin{subfigure}[t]{0.25\textwidth}
        \centering
        \includegraphics[width=\linewidth, height=.7\linewidth]{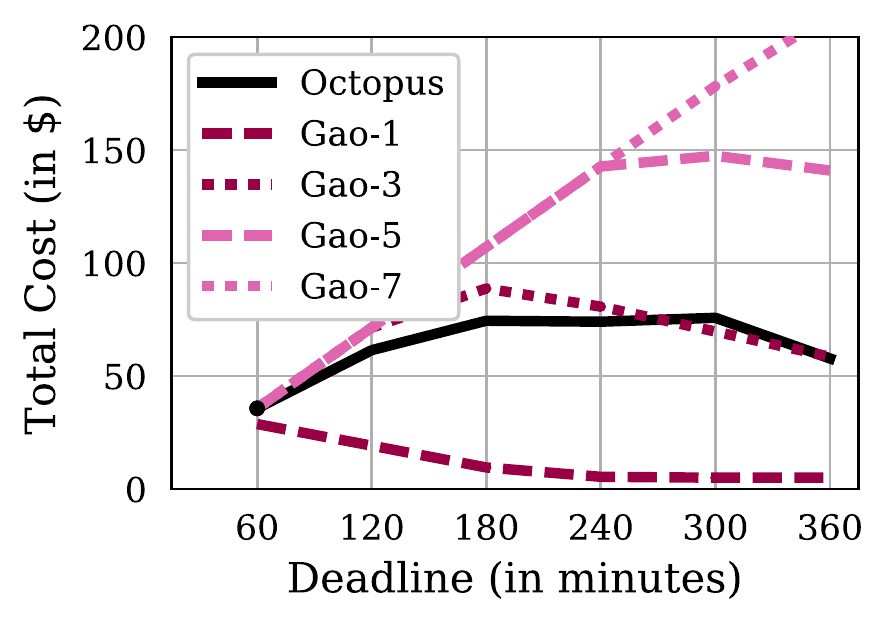}
        \captionsetup{width=0.93\linewidth}
        \caption{Comparison to {\sc Gao}.}
        \label{fig:controller_live}
    \end{subfigure}
    \caption{Performance of \octo\ on simulated data for several deadlines. (a), (c) are plots of the utility achieved. Each data point in (b) represents the performance averaged over all deadlines considered in (a). (d), (f) compare the accuracy achieved while (e), (g) compare the total cost incurred by \octo\ and competing baselines. Round dots on the figures denote that the difference between \oct \ and the competing baseline (whether better or worse) is statistically insignificant at that point. All other differences are statistically significant.}

    \label{fig:simulations}
\end{figure*}

\begin{figure}[b!!]
    \centering
    \begin{subfigure}{0.24\textwidth}
        \centering
        \includegraphics[width=\linewidth]{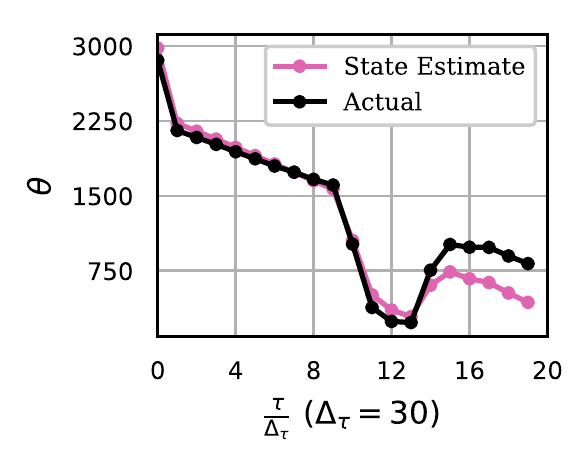}
        \label{fig:tracking_bhat}
    \end{subfigure}%
    \begin{subfigure}{0.24\textwidth}
        \centering
        \includegraphics[width=\linewidth]{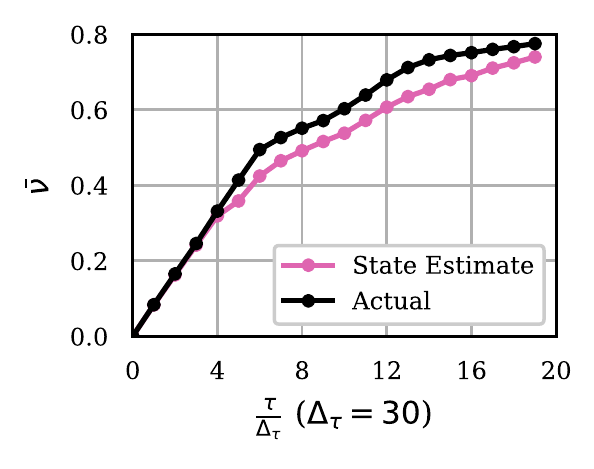}
        \label{fig:tracking_vbar}
    \end{subfigure}
    \caption{Tracking the real state using $\beta$-reconstruction in the \cost\ for the (a) $\theta$ value, and (b) $\bar{\nu}$ value.}
    \label{fig:unsync_tracking}
\end{figure}

\begin{figure}[t]
    \centering
    \begin{subfigure}{0.24\textwidth}
        \centering
        \includegraphics[width=\linewidth]{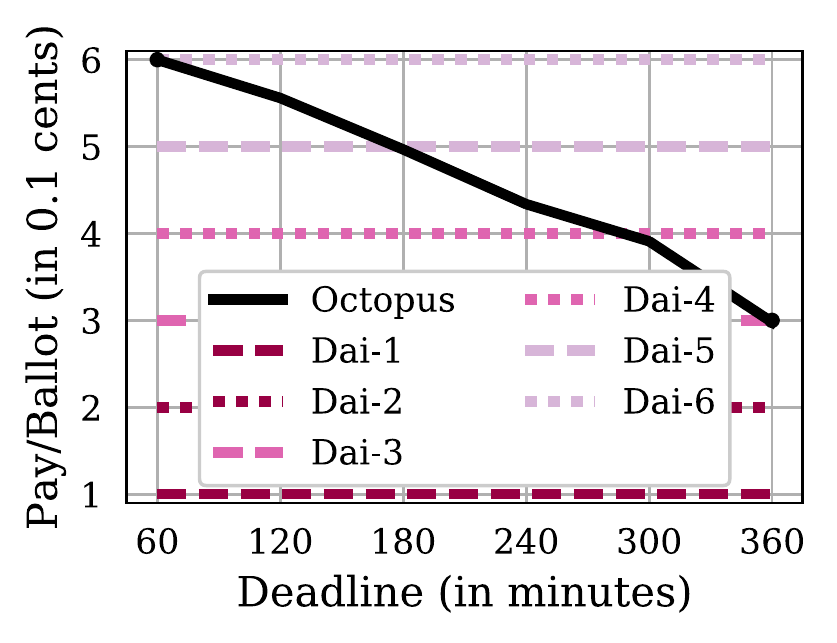}
        \label{fig:payperballotdai}
    \end{subfigure}%
    \begin{subfigure}{0.24\textwidth}
        \centering
        \includegraphics[width=\linewidth]{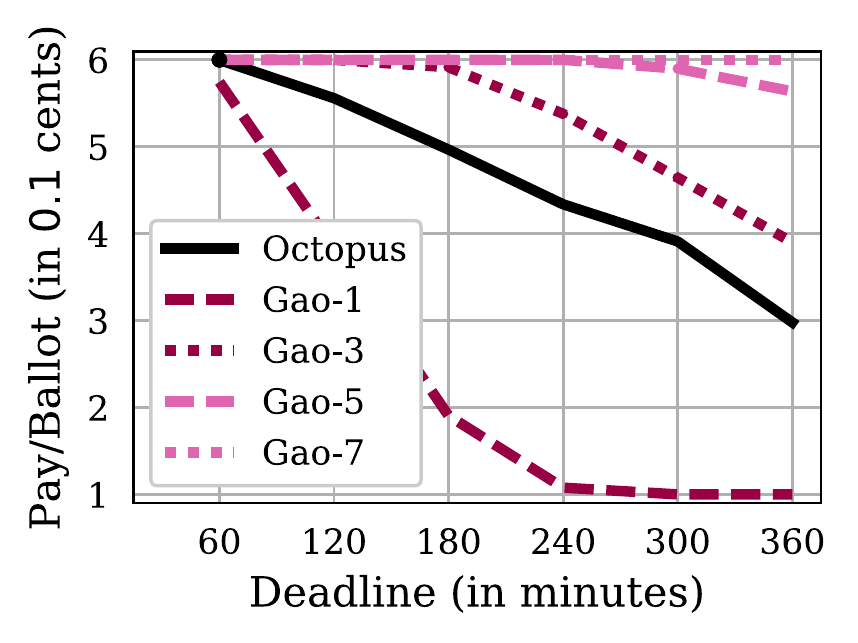}
        \label{fig:payperballotgao}
    \end{subfigure}
    \caption{Comparison of the average pay/ballot against (a) {\sc Dai}, and (b) {\sc Gao} on simulated data for several deadlines. }
    \label{fig:payperballot}
\end{figure}

\begin{figure*}[t!]
	\centering
    \begin{subfigure}[t]{0.3\textwidth}
        \centering              
        \captionsetup{width=0.93\linewidth}
        \includegraphics[width=\linewidth, height=.7\linewidth]{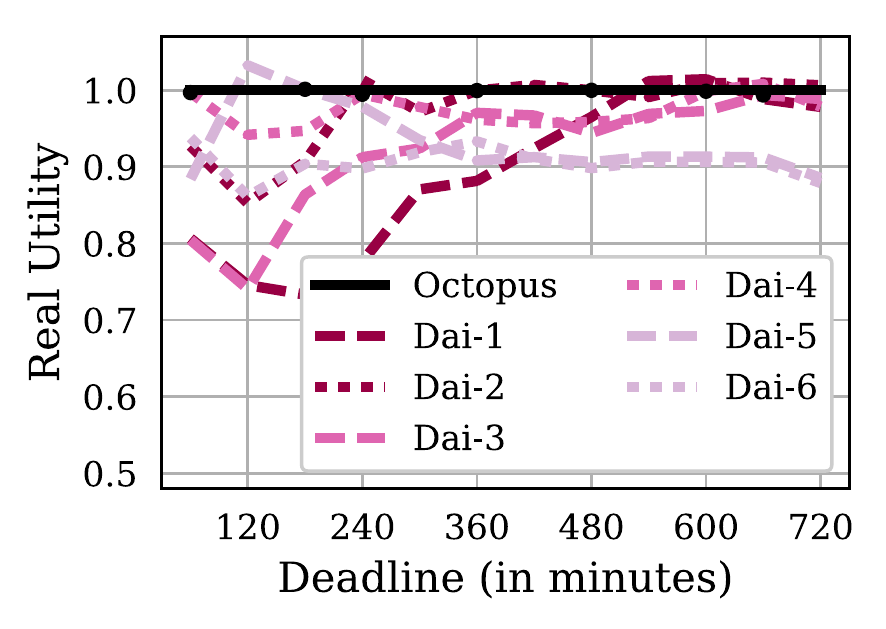}
        \caption{Comparison to \textsc{Dai}.}
        \label{fig:static_live}
    \end{subfigure}
    \begin{subfigure}[t]{0.3\textwidth}
        \centering
        \includegraphics[width=\linewidth, height=.7\linewidth]{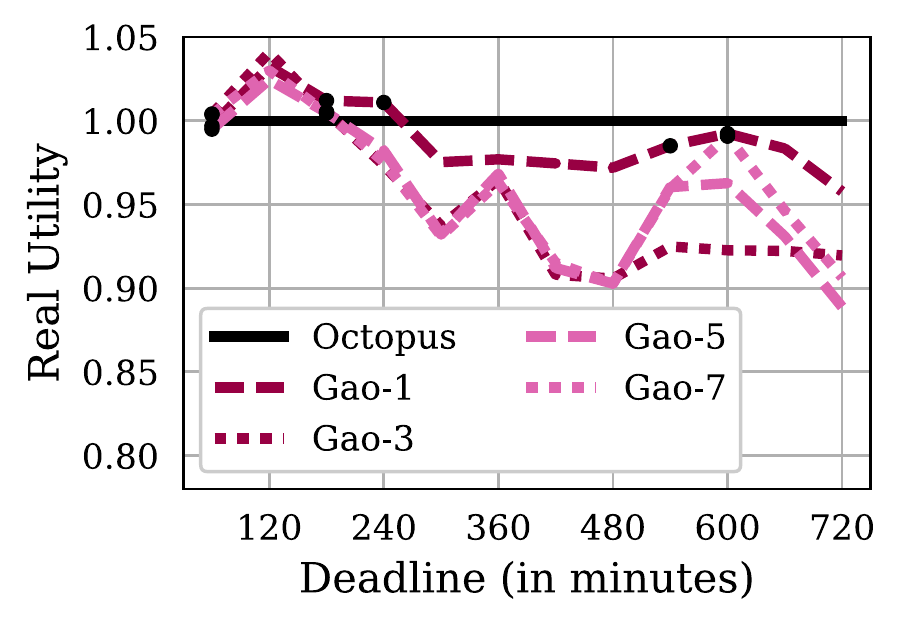}
        \captionsetup{width=0.93\linewidth}
        \caption{Comparison to \textsc{Gao}.}
        \label{fig:gao_live}
    \end{subfigure}%
        \begin{subfigure}[t]{0.3\textwidth}
        \centering
        \includegraphics[width=\linewidth, height=.7\linewidth]{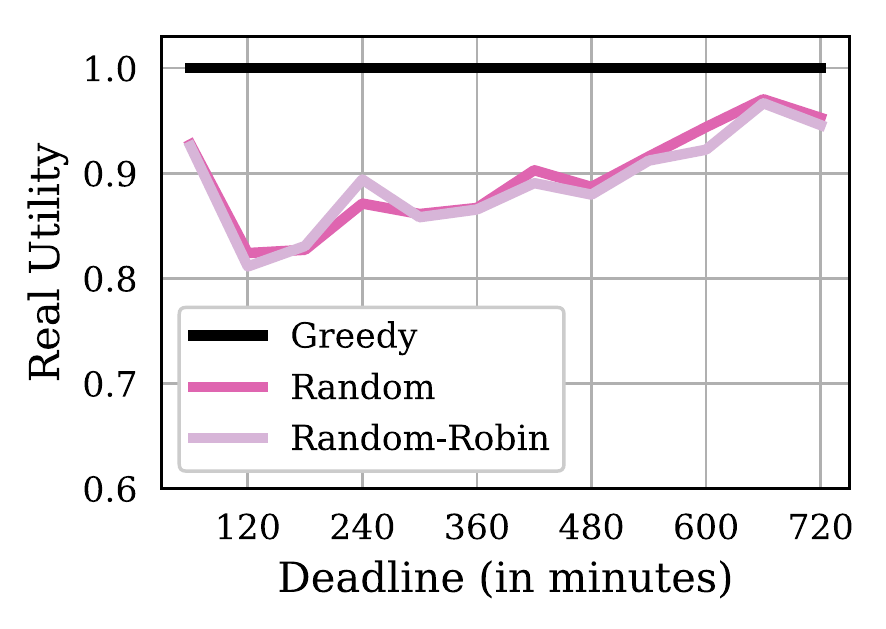}
        \captionsetup{width=0.93\linewidth}
        \caption{Advantage of {\sc Greedy}.}
        \label{fig:controller_live}
    \end{subfigure}
    
    \begin{subfigure}[t]{0.25\textwidth}
        \centering              
        \captionsetup{width=0.93\linewidth}
        \includegraphics[width=\linewidth, height=.7\linewidth]{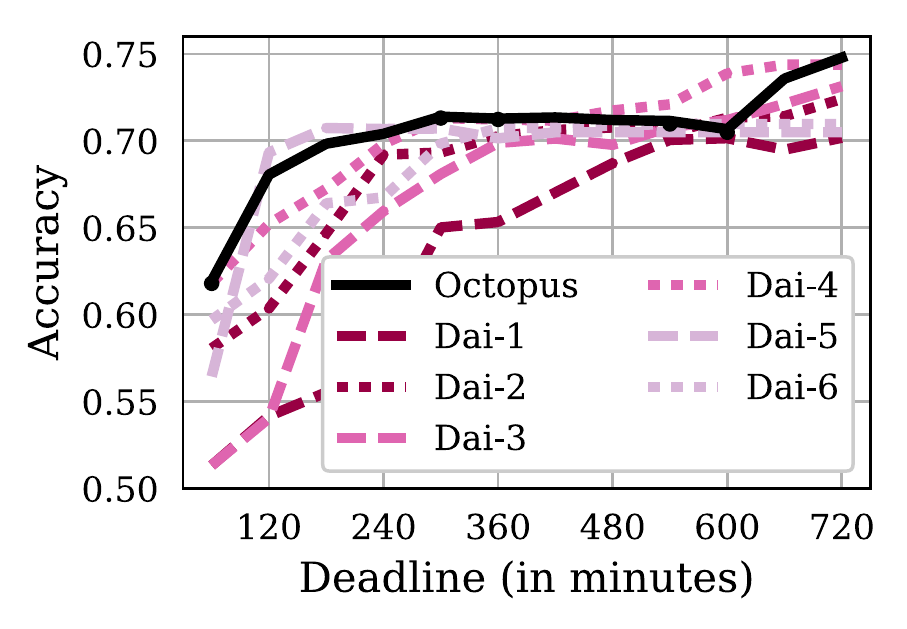}
        \caption{Comparison to \textsc{Dai}.}
        \label{fig:static_live}
    \end{subfigure}%
    \begin{subfigure}[t]{0.25\textwidth}
        \centering
        \includegraphics[width=\linewidth, height=.7\linewidth]{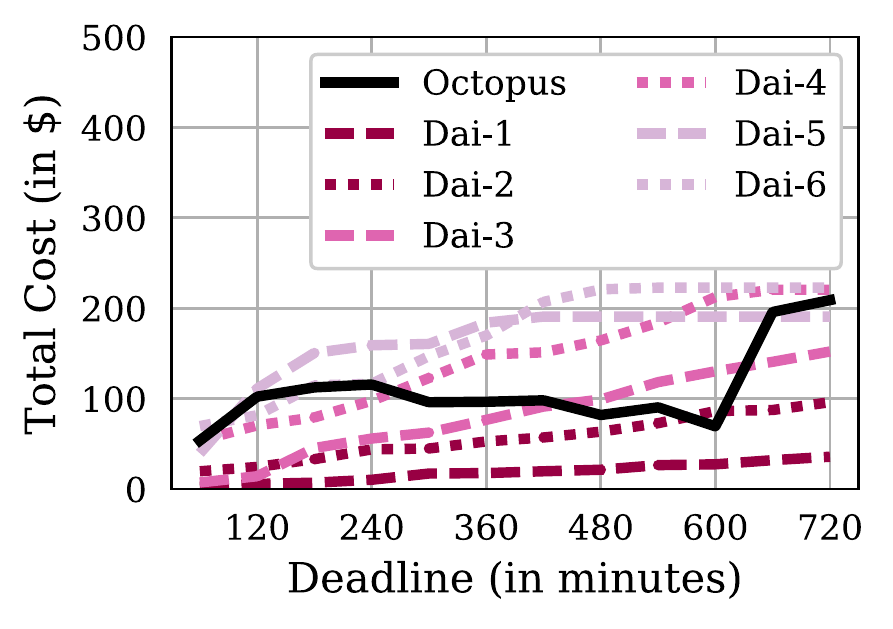}
        \captionsetup{width=0.93\linewidth}
        \caption{Comparison to \textsc{Dai}.}
        \label{fig:gao_live}
    \end{subfigure}%
        \begin{subfigure}[t]{0.25\textwidth}
        \centering
        \includegraphics[width=\linewidth, height=.7\linewidth]{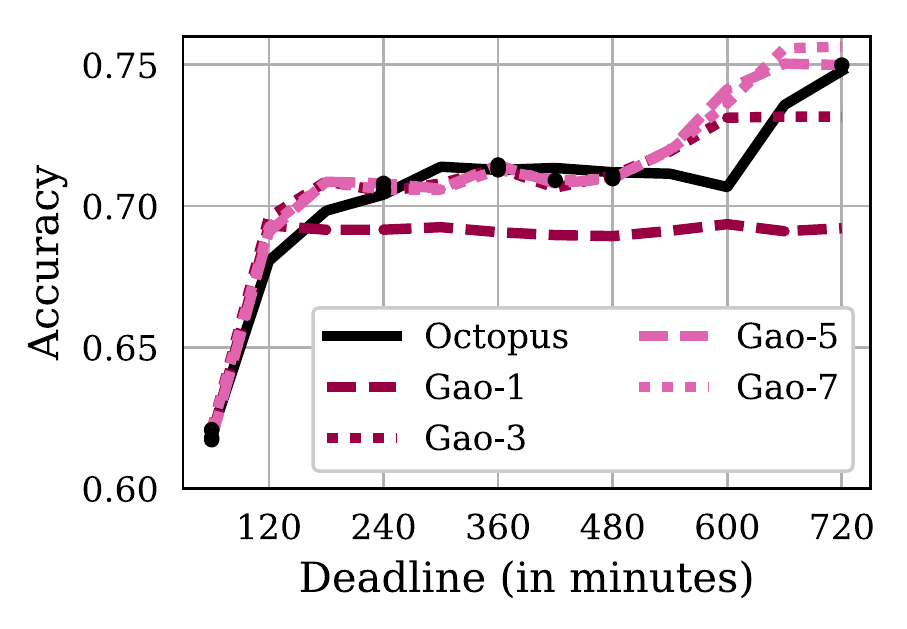}
        \captionsetup{width=0.93\linewidth}
        \caption{Comparison to {\sc Gao}.}
        \label{fig:controller_live}
    \end{subfigure}%
    \begin{subfigure}[t]{0.25\textwidth}
        \centering
        \includegraphics[width=\linewidth, height=.7\linewidth]{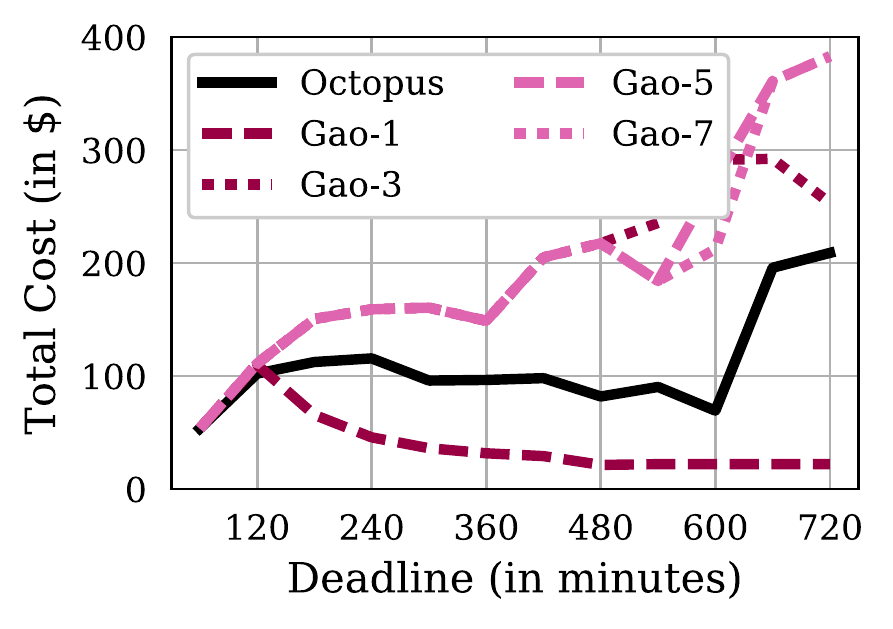}
        \captionsetup{width=0.93\linewidth}
        \caption{Comparison to {\sc Gao}.}
        \label{fig:controller_live}
    \end{subfigure}
    \caption{Performance of \octo\ on real data for several deadlines. (a), (c) are plots of the utility achieved. Each data point in (b) represents the performance averaged over all deadlines considered in (a). (d), (f) compare the accuracy achieved while (e), (g) compare the total cost incurred by \octo\ and competing baselines. Round dots on the figures denote that the difference between \oct \ and the competing baseline (whether better or worse) is statistically insignificant at that point. All other differences are statistically significant.} 
    \label{fig:real_data}
\end{figure*}

\subsection{Simulation Experiments}
Our main aim through simulations is to assess the quality of $\beta$-reconstruction. We use a variety of parameter settings and simulate \octo\ with ballot arrivals simulated according to a pay-dependent Poisson process, and answers generated based on worker models. Note that this experiment is {\em without} any synchronization between the \cost\ state and the real batch. Figure~\ref{fig:unsync_tracking}a compares the $\bhat$ values between the system and the batch (for one such setting\footnote{$500$ tasks with $p(d) =\beta(2.0,2.0)$, worker errors sampled from $\Gamma(2.0,0.5)$, ${\cal P} = 200$, $\Delta_\tau = 15$ mins, $\Delta_{\theta} = 10$, $\Delta_{\vbar} = 100$.}). Even after 10 time intervals, and across multiple cost changes (the sharp drops in the curve), our tracking is extremely accurate. Figure~\ref{fig:unsync_tracking}b shows that our $\vbar$ tracking is extremely effective as well, with very little divergence from the true value. Even when the values do diverge in both cases, they are highly correlated. 

High quality tracking of $\bhat$ and $\vbar$ are essential\shorten{, since the \cost\ optimizes long-term expected reward}. If we diverge too far from the real values, we would likely optimize the long-term expected reward poorly. Overall, this reaffirms the hypothesis that the \cost\ is able to capture the global state of the whole batch using just the aggregate statistics.

\shorten{(b) assess its impact on performance by using \octo\ without synchronization; (c) verify robustness to parameter changes.}

Our other goal is to compare \octo's performance with baselines in simulation, which we do below. 


\noindent 
{{\bf Comparison to {\sc Dai}}. \shorten{We first address the efficacy of our \cost's behavior.} Comparing \oct\ with {\sc Dai} \shorten{allows us to determine whether} highlights the benefit of changing cost \shorten{can have a pronounced effect}on real-world utility. {Ideally, our method should be able to vary pay to match or exceed the utility of the static cost baselines. } We run both \oct\ and {\sc Dai} with deadlines ranging from 60 to 360 minutes. We run {\sc Dai} for different static pays, ranging from 1 to 6 {(measured in a tenth of a cent)}, allowing it to take ballots until the deadline. Statistical significance is indicated on the plots.

Figure~\ref{fig:simulations} shows the comparison. \oct\ simultaneously outperforms {(or is at par with)} all static cost baselines for \emph{every} deadline, whereas each static cost baseline has a `sweet spot' range of deadlines where it does best.\shorten{the higher cost baselines only do well for shorter deadlines (\emph{e.g.} cost 6 for 1 hr), and the lower cost baselines only perform well given longer deadlines (\emph{e.g.} cost 1 for 6 hrs).} \oct\ plans robustly, where despite making pricing decisions 24 times without synchronizing for the $6$ hr deadline, it still outperforms {\sc Dai}. \shorten{This makes sense intuitively, since \oct\ is able to plan for different arrival rates.  {\sc Dai} is unable to plan for different arrival rates. By changing pay, \oct\ can decrease the effective rate paid over time, while still achieving high accuracy}\shorten{\oct\ is able to change cost extremely effectively, and is almost $50\%$ better than cost $1$ in the short 1 hr deadline, and around $65\%$ better than cost $6$ in the 6 hr deadline.} Figure~\ref{fig:simulations}b depicts the utility averaged across all deadlines as a function of task-completion rates. As shown, \oct\ is robust to different task-completion rates, and continues to outperform {\sc Dai} on changing them. 
Figure~\ref{fig:simulations}d \& \ref{fig:simulations}e explain why \octo\ achieves better utility scores than {\sc Dai} -- it maintains very high accuracy while keeping costs reasonable. Qualitatively, for every deadline, we observe that \octo\ tends to have an average cost that is close to the best static cost.

\noindent
{\bf Comparison to {\sc Gao}}. Comparing with {\sc Gao} allows us to delineate the effect of the \quality s while optimizing for batch quality using aggregate statistics. 

For fairness, we augment {\sc Gao}'s framework; fixing $r$ ballots/task up-front, 
and using the same worker response model as \oct\ 
for ballot aggregation. 
Figure~\ref{fig:simulations}c demonstrates \oct's performance against {\sc Gao} for different values of $r$. \oct\ consistently outperforms {\sc Gao} for all values of $r$ and across all deadlines. \oct's performance improves for longer deadlines, due to the higher quality achieved by the \quality. For instance, \oct\ is around $100\%$ better than {\sc Gao} for the 6 hr deadline. We see that {\sc Gao}-3 and \octo\ incur nearly the same cost (Figure~\ref{fig:simulations}g) but \octo\ is $\sim$6\% better in terms of accuracy (Figure~\ref{fig:simulations}f). Examining Figure~\ref{fig:payperballot}b reveals that this is in part due to \octo\ collecting $\times \frac{1}{3}$ \emph{more} ballots than {\sc Gao}-3 by changing pay/ballot more intelligently. 

It is interesting to note that increasing $r$ does not significantly improve baseline performance. This is due to the fact that the \quality\ only takes more ballots on tasks that really need them. For {\sc Gao} the extra cost spent on every task is not offset by a corresponding increase in quality, especially for tasks that are very easy (don't require $r$ ballots) or too hard (unsolved even with $r$ ballots).}
In Figure~\ref{fig:simulations}f \&  \ref{fig:simulations}g, we notice that \octo\ once again has near-highest accuracy, but spends much less than {\sc Gao} variants that have higher accuracy than \octo.

Lastly, Figure~\ref{fig:payperballot} demonstrates that \octo\ has intuitive behavior -- as the deadline length increases, \octo\ spends less pay/ballot on average since more time can be taken to finish the batch of tasks.

These experiments demonstrate that \octo\ \emph{simultaneously} achieves higher utility than every variant of baseline methods, and does so without sacrificing accuracy or incurring high costs. 

\noindent

\subsection{Offline Experiments on Real Data}

We run extensive offline experiments on our collected data against the state-of-the-art baselines described earlier.

\shorten{We first learn average arrival rates at each price point from the data -- useful for \cost's MDP.} When comparing algorithms we sample different execution trajectories from the real data by choosing a random ballot for each task (since quality is cost-independent), while keeping ballot arrivals as per the real data. Multiple trajectories help compute statistical significance over algorithms' performances. These experiments synchronize the \cost's state with the real batch after every $\Delta_\tau$ time. We use an uninformed uniform distribution for the difficulty prior  $p(d)$.

Figure~\ref{fig:real_data}a shows \octo\ when compared to {\sc Dai} at different static price points, with the $x$-axis being different $\tau_{\max}$ values. Statistical significance is marked on the plots. \octo\ outperforms most {\sc Dai} costs across all deadlines, with upto $37\%$ increase in real utility.  No single static cost is able to match \octo\ across all deadlines. This underscores the benefits of changing pay dynamically based on current task completion. In Figure~\ref{fig:real_data}d \& \ref{fig:real_data}e, we see that \octo\ achieves high accuracy at relatively low cost.

Figure~\ref{fig:real_data}b compares \octo\ with {\sc Gao}-$r$, with $r$ denoting the static number of ballots per task. 
We find that \octo\ outperforms {\sc Gao} for most deadlines.  Further analysis reveals that our batch exhibits a bi-modal task difficulty distribution -- no algorithm exceeds around $76\%$ accuracy, while getting $70\%$ is easy with $\sim$1 ballot per task (in Figure~\ref{fig:real_data}f, we see a sharp jump in accuracy only when the deadline is long enough to take a lot of ballots). For shorter deadlines, where  solving difficult tasks is infeasible, {\sc Gao} outperforms \octo\ slightly, since \octo\ optimizes with respect to a uniform prior. For longer deadlines, estimates of task difficulties are refined by the \quality s, and \octo\ \shorten{synchronizes its state and} gives large gains.

Lastly, \octo\ outperforms \controller\ baselines where {\sc Greedy} task selection is replaced by a random policy ({\sc Random}), or a single round robin followed by random selection ({\sc Random-Robin}), by large margins (Figure \ref{fig:real_data}c). Note also that the baselines start to converge over time, as the advantage of doing intelligent task selection diminishes when nearly all tasks are run till completion.

Overall, we find that \octo\ learns robust policies, consistently outperforming all baselines.

\subsection{Live Online Experiments}

Lastly, we deploy \octo\ on AMT to test performance in a dynamic, online setting, as well as gain qualitative insight into worker behavior. In this, we keep exactly 3 HITS (of 10 tasks each) on AMT at a time. If a worker accepts a HIT, another one is posted immediately -- this enables greedy task routing while ensuring full power of worker parallelism. After every $\Delta_\tau$ mins, all available HITs are taken down and reposted with the new price output by \octo. \octo's policies are learned using task completion rates estimated from real data earlier. At runtime, querying \octo\ is instantaneous. We solve a batch of 500 tweets for 3 deadlines -- 1, 2, and 4 hours. We compare against {\sc Gao}-$1$, the best baseline in offline expts in Table~\ref{tab:live_results}.


For the 1 hr deadline, \octo\ maintains pay at 0.5 cent/ballot, before decreasing it to 0.1 cent/ballot in the last 15 minutes. This allows \octo\ to receive around 1 ballot/task; more ballots stagnate ${\cal U}$ for short deadlines due to conflicting workers, and \octo\ prefers to save money to optimize utility. On the other hand, {\sc Gao} maintains pay at 0.6 cent throughout to also ensure it receives 1 ballot/task, but is unlucky in the responses it receives. In terms of decision making, the superiority of \octo\ is clear, since it recognizes the danger posed by disagreement on task quality estimates. 

In the 2 hr deadline, \octo\ once again increases pay initially, but gets more ballot arrivals than expected. In response, after 45 minutes \octo\ decreases pay so that it can take a higher number of ballots for difficult tasks, getting a substantial accuracy and utility improvement over {\sc Gao}.


For the 4 hr deadline, \octo\ is aggressive in trying to solve all tasks till completion. Due to the bi-modal difficulty of the tasks, workers provide several conflicting ballots on harder tasks, which the \controller\ prefers to re-route for utility gain. The overall accuracy increases marginally over the 2 hour deadline. 

\begin{table}[t!]
{\small
\centering
\begin{tabular}{ccccccc}
\toprule
\multicolumn{1}{c}{Method} & \multicolumn{3}{c}{{\sc Octopus}} & \multicolumn{3}{c}{{\sc Gao} $(r = 1)$} \\ \toprule
Deadline                     & Cost       & Acc.     & ${\cal U}$         & Cost         & Acc.       & ${\cal U}$           \\ 
1 hr                         & \$2.63     & 74.4\%        & {\bf -25.9}    & \$2.68       & 71.4\%          & -28.9      \\ 
2 hrs                        & \$3.17     & 77.2\%        & {\bf -23.1}    & \$1.72       & 73.8\%          & -26.4      \\ 
4 hrs                        & \$6.23     & 77.4\%        & {\bf -23.2}    & \$0.50       & 74.4\%          & -25.6      \\ \hline
\end{tabular}
}
\caption{Online performance comparison of \octo\ and {\sc Gao} ($r = 1$) for a single trial. ${\cal U}$ is shown in thousands.}
\label{tab:live_results}
\end{table}

Qualitatively, workers respond as predicted -- flocking to the tasks when pay was set at 0.5 cent or more, and staying away at very low pay. Workers respond naturally to the price changing algorithm; dropping out immediately if the pay is suddenly lowered, and coming back if it is increased once again. No worker complained about the fluctuating pay.

\subsection{Discussion}
 {In this work, we focused on experiments with \octo\ in a fixed time deadline setting to compare with past work. Extension to the popular fixed budget setting (where $\sum_q C_q$ is constrained) is simple: (i) never synchronize $\theta$ so that its value in any state is simply the start value of $\theta$ (which is fixed and known) minus the ballots received, making it a proxy for cost incurred so far; (ii) modify the \cost's reward to give a $-\infty$ reward for total cost (now computable using $\theta$) exceeding the budget. Another extension involves using a different formulation of the \quality; any algorithm that defines an appropriate policy and from which we can compute $\theta$ and $\bar{\nu}$ is suitable.}

\section{Conclusion}
We present \octo, one of the first AI agents for a 3-way optimization of total cost, work quality and completion time in crowdsourcing. The agent combines three different subagents that control quality per task, select the best next task and set pay for the whole batch. A key technical contribution is the computation of aggregate statistics of the quality and completeness of the whole batch -- this is used as the state for best setting the next pay.

\octo\ outperforms state-of-the-art baselines in a variety of simulated and real world settings, demonstrating the superiority of our approach. We also showcase \octo 's real world applicability by deploying it directly on AMT.  In the future, we hope to develop general purpose formulations of \octo\ as a plug-and-play architecture for practitioners. 

\section{Acknowledgments}
This work is supported by Google language understanding and knowledge discovery focused research grants, a Bloomberg award, a Microsoft Azure sponsorship, and a Visvesvaraya faculty award by Govt. of India to the third author. We thank Chris Lin, Yihan Gao and Aditya Parameswaran for sharing code, and all the AMT workers who participated in our experiments. 


\fontsize{9.0pt}{10.0pt}
\bibliographystyle{aaai} 
\bibliography{bibfile}

\begin{thebibliography}{}

\bibitem[\protect\citeauthoryear{Ambati, Vogel, and
  Carbonell}{2011}]{ambati2011towards}
Ambati, V.; Vogel, S.; and Carbonell, J.~G.
\newblock 2011.
\newblock Towards task recommendation in micro-task markets.
\newblock In {\em {AAAI} Workshop on Human Computation}.

\bibitem[\protect\citeauthoryear{Bragg \bgroup et al\mbox.\egroup
  }{2014}]{bragg2014parallel}
Bragg, J.; Kolobov, A.; Mausam; and Weld, D.~S.
\newblock 2014.
\newblock Parallel task routing for crowdsourcing.
\newblock In {\em HCOMP}.

\bibitem[\protect\citeauthoryear{Bragg, Mausam, and
  Weld}{2013}]{bragg2013crowdsourcing}
Bragg, J.; Mausam; and Weld, D.~S.
\newblock 2013.
\newblock Crowdsourcing multi-label classification for taxonomy creation.
\newblock In {\em HCOMP}.

\bibitem[\protect\citeauthoryear{Chatterjee, Majumdar, and
  Henzinger}{2006}]{chatterjee2006markov}
Chatterjee, K.; Majumdar, R.; and Henzinger, T.~A.
\newblock 2006.
\newblock Markov decision processes with multiple objectives.
\newblock In {\em STACS 2006}. Springer.

\bibitem[\protect\citeauthoryear{Dai \bgroup et al\mbox.\egroup
  }{2013}]{dai2013pomdp}
Dai, P.; Lin, C.~H.; Mausam; and Weld, D.~S.
\newblock 2013.
\newblock Pomdp-based control of workflows for crowdsourcing.
\newblock {\em Artif. Intell.} 202:52--85.

\bibitem[\protect\citeauthoryear{Dai \bgroup et al\mbox.\egroup
  }{2015}]{dai2015and}
Dai, P.; Rzeszotarski, J.~M.; Paritosh, P.; and Chi, E.~H.
\newblock 2015.
\newblock And now for something completely different: Improving crowdsourcing
  workflows with micro-diversions.
\newblock In {\em CSCW}.
\newblock ACM.

\bibitem[\protect\citeauthoryear{Dai, Mausam, and
  Weld}{2011}]{Dai2011ArtificialIF}
Dai, P.; Mausam; and Weld, D.~S.
\newblock 2011.
\newblock Artificial intelligence for artificial artificial intelligence.
\newblock In {\em AAAI}.

\bibitem[\protect\citeauthoryear{Difallah \bgroup et al\mbox.\egroup
  }{2014}]{difallah2014scaling}
Difallah, D.~E.; Catasta, M.; Demartini, G.; and Cudr{\'e}-Mauroux, P.
\newblock 2014.
\newblock Scaling-up the crowd: Micro-task pricing schemes for worker retention
  and latency improvement.
\newblock In {\em HCOMP}.

\bibitem[\protect\citeauthoryear{Faradani, Hartmann, and
  Ipeirotis}{2011}]{faradani2011s}
Faradani, S.; Hartmann, B.; and Ipeirotis, P.~G.
\newblock 2011.
\newblock What's the right price? pricing tasks for finishing on time.
\newblock In {\em {AAAI} Workshop on Human Computation}.

\bibitem[\protect\citeauthoryear{Gao and Parameswaran}{2014}]{gao2014finish}
Gao, Y., and Parameswaran, A.~G.
\newblock 2014.
\newblock Finish them!: Pricing algorithms for human computation.
\newblock {\em {PVLDB}}.

\bibitem[\protect\citeauthoryear{Golovin and
  Krause}{2011}]{golovin2011adaptive}
Golovin, D., and Krause, A.
\newblock 2011.
\newblock Adaptive submodularity: Theory and applications in active learning
  and stochastic optimization.
\newblock {\em Journal of Artificial Intelligence Research}.

\bibitem[\protect\citeauthoryear{Haas \bgroup et al\mbox.\egroup
  }{2015}]{haas2015clamshell}
Haas, D.; Wang, J.; Wu, E.; and Franklin, M.~J.
\newblock 2015.
\newblock Clamshell: speeding up crowds for low-latency data labeling.
\newblock {\em VLDB}.

\bibitem[\protect\citeauthoryear{Hansen and
  Zilberstein}{2001}]{hansen2001monitoring}
Hansen, E.~A., and Zilberstein, S.
\newblock 2001.
\newblock Monitoring and control of anytime algorithms: A dynamic programming
  approach.
\newblock {\em Artificial Intelligence} 126(1-2):139--157.

\bibitem[\protect\citeauthoryear{Ipeirotis and
  Gabrilovich}{2014}]{ipeirotis2014quizz}
Ipeirotis, P.~G., and Gabrilovich, E.
\newblock 2014.
\newblock Quizz: Targeted crowdsourcing with a billion (potential) users.
\newblock In {\em WWW}.
\newblock ACM.

\bibitem[\protect\citeauthoryear{Kamar \bgroup et al\mbox.\egroup
  }{2013}]{kamar2013lifelong}
Kamar, E.; Kapoor, A.; Horvitz, E.; and Redmond, W.
\newblock 2013.
\newblock Lifelong learning for acquiring the wisdom of the crowd.
\newblock In {\em IJCAI}.
\newblock Citeseer.

\bibitem[\protect\citeauthoryear{Karger, Oh, and Shah}{2014}]{karger2014budget}
Karger, D.~R.; Oh, S.; and Shah, D.
\newblock 2014.
\newblock Budget-optimal task allocation for reliable crowdsourcing systems.
\newblock {\em Operations Research}.

\bibitem[\protect\citeauthoryear{Kearns, Mansour, and
  Ng}{1999}]{kearns1999approximate}
Kearns, M.~J.; Mansour, Y.; and Ng, A.~Y.
\newblock 1999.
\newblock Approximate planning in large pomdps via reusable trajectories.
\newblock Citeseer.

\bibitem[\protect\citeauthoryear{Kobren \bgroup et al\mbox.\egroup
  }{2015}]{kobren2015getting}
Kobren, A.; Tan, C.~H.; Ipeirotis, P.; and Gabrilovich, E.
\newblock 2015.
\newblock Getting more for less: optimized crowdsourcing with dynamic tasks and
  goals.
\newblock In {\em WWW}.

\bibitem[\protect\citeauthoryear{Lin, Mausam, and
  Weld}{2012}]{Lin2012CrowdsourcingCM}
Lin, C.~H.; Mausam; and Weld, D.~S.
\newblock 2012.
\newblock Crowdsourcing control: Moving beyond multiple choice.
\newblock In {\em UAI}.

\bibitem[\protect\citeauthoryear{Mason and Watts}{2010}]{mason2010financial}
Mason, W., and Watts, D.~J.
\newblock 2010.
\newblock Financial incentives and the performance of crowds.
\newblock {\em ACM SigKDD Explorations Newsletter} 11(2):100--108.

\bibitem[\protect\citeauthoryear{Oleson \bgroup et al\mbox.\egroup
  }{2011}]{oleson2011programmatic}
Oleson, D.; Sorokin, A.; Laughlin, G.~P.; Hester, V.; Le, J.; and Biewald, L.
\newblock 2011.
\newblock Programmatic gold: Targeted and scalable quality assurance in
  crowdsourcing.
\newblock {\em Human computation}.

\bibitem[\protect\citeauthoryear{Parameswaran \bgroup et al\mbox.\egroup
  }{2012}]{parameswaran2012crowdscreen}
Parameswaran, A.~G.; Garcia{-}Molina, H.; Park, H.; Polyzotis, N.; Ramesh, A.;
  and Widom, J.
\newblock 2012.
\newblock Crowdscreen: algorithms for filtering data with humans.
\newblock In {\em SIGMOD}.

\bibitem[\protect\citeauthoryear{Rajpal, Goel, and Mausam}{2015}]{rajpalpomdp}
Rajpal, S.; Goel, K.; and Mausam.
\newblock 2015.
\newblock Pomdp-based worker pool selection for crowdsourcing.
\newblock {\em CrowdML Workshop, ICML}.

\bibitem[\protect\citeauthoryear{Rzeszotarski \bgroup et al\mbox.\egroup
  }{2013}]{rzeszotarski2013inserting}
Rzeszotarski, J.~M.; Chi, E.; Paritosh, P.; and Dai, P.
\newblock 2013.
\newblock Inserting micro-breaks into crowdsourcing workflows.
\newblock In {\em HCOMP}.

\bibitem[\protect\citeauthoryear{Shahaf and
  Horvitz}{2010}]{shahaf2010generalized}
Shahaf, D., and Horvitz, E.
\newblock 2010.
\newblock Generalized task markets for human and machine computation.
\newblock In {\em AAAI}.

\bibitem[\protect\citeauthoryear{Sheshadri and
  Lease}{2013}]{Sheshadri2013SQUAREAB}
Sheshadri, A., and Lease, M.
\newblock 2013.
\newblock Square: A benchmark for research on computing crowd consensus.
\newblock In {\em HCOMP}.

\bibitem[\protect\citeauthoryear{Smith and Simmons}{2012}]{smith2012point}
Smith, T., and Simmons, R.
\newblock 2012.
\newblock Point-based pomdp algorithms: Improved analysis and implementation.
\newblock {\em arXiv preprint arXiv:1207.1412}.

\bibitem[\protect\citeauthoryear{Teneketzis and
  Ho}{1987}]{teneketzis1987decentralized}
Teneketzis, D., and Ho, Y.-C.
\newblock 1987.
\newblock The decentralized wald problem.
\newblock {\em Information and Computation} 73(1):23--44.

\bibitem[\protect\citeauthoryear{Venetis \bgroup et al\mbox.\egroup
  }{2012}]{venetis2012max}
Venetis, P.; Garcia-Molina, H.; Huang, K.; and Polyzotis, N.
\newblock 2012.
\newblock Max algorithms in crowdsourcing environments.
\newblock In {\em Proceedings of the 21st international conference on World
  Wide Web},  989--998.
\newblock ACM.

\bibitem[\protect\citeauthoryear{Weld \bgroup et al\mbox.\egroup
  }{2015}]{weld2015handbook}
Weld, D.~S.; Mausam; Lin, C.~H.; and Bragg, J.
\newblock 2015.
\newblock Artificial intelligence and collective intelligence.
\newblock {\em Handbook of Collective Intelligence}.

\bibitem[\protect\citeauthoryear{Welinder and
  Perona}{2010}]{welinder2010online}
Welinder, P., and Perona, P.
\newblock 2010.
\newblock Online crowdsourcing: rating annotators and obtaining cost-effective
  labels.

\bibitem[\protect\citeauthoryear{Welinder \bgroup et al\mbox.\egroup
  }{2010}]{welinder2010multidimensional}
Welinder, P.; Branson, S.; Perona, P.; and Belongie, S.~J.
\newblock 2010.
\newblock The multidimensional wisdom of crowds.
\newblock In {\em NIPS}.

\bibitem[\protect\citeauthoryear{Whitehill \bgroup et al\mbox.\egroup
  }{2009}]{whitehill2009whose}
Whitehill, J.; Wu, T.-f.; Bergsma, J.; Movellan, J.~R.; and Ruvolo, P.~L.
\newblock 2009.
\newblock Whose vote should count more: Optimal integration of labels from
  labelers of unknown expertise.
\newblock In {\em NIPS}.

\end{thebibliography}

\end{document}